\documentclass[sigconf,nonacm]{acmart}
\usepackage{dblfloatfix} % 修复双栏浮动体问题
\usepackage{graphicx}
\usepackage{multirow}    % 多行合并
\usepackage{wrapfig} % 可选，用于文字环绕
\usepackage[table]{xcolor}
\usepackage{float}
\usepackage{rotating} % 提供 sideways 环境
\usepackage{booktabs} % ACM 表格必备（无竖线）
\usepackage{makecell} % 优化表头换行
\usepackage{hyphenat}         % 核心断字包
\usepackage{subcaption} % 子图

\usepackage{seqsplit}         % 处理超长单词/代码
\usepackage{ragged2e}         % 改进文本对齐
\usepackage{enumitem}
\usepackage{xcolor}
\apptocmd{\sloppy}{\hbadness=10000\relax}{}{}

\AtBeginDocument{%
  }

\setcopyright{none} 
\acmConference[Anonymous Submission]{Anonymous Submission}
\acmYear{2026}
\acmDOI{} % 保留空DOI
\acmISBN{} % 保留空ISBN

\begin{document}

\title{MEMTS: Internalizing Domain Knowledge via Parameterized Memory for Retrieval-Free Domain Adaptation of Time Series Foundation Models}


\author{Xiaoyun Yu}
\authornote{Both authors contributed equally to this research.} % 1. 定义共同一作说明
\affiliation{%
  \institution{East China Normal University}
  \city{Shanghai}
  \country{China}}
\email{xyyu@stu.ecnu.edu.cn}

\author{Li Fan}
\authornotemark[1] % 2. 复用第1个脚注的符号（通常是 * 或 †）
\affiliation{%
  \institution{East China Normal University}
  \city{Shanghai}
  \country{China}}
\email{lfan@stu.ecnu.edu.cn}

\author{Xiangfei Qiu}
\affiliation{%
  \institution{East China Normal University}
  \city{Shanghai}
  \country{China}}
\email{xfqiu@stu.ecnu.edu.cn}

\author{Nanqing Dong}
\affiliation{%
  \institution{Shanghai Artificial Intelligence Laboratory}
  \city{Shanghai}
  \country{China}}
\email{dongnanqing@pjlab.org.cn}

\author{Yonggui Huang}
\affiliation{%
  \institution{Peking University}
  \city{Beijing}
  \country{China}}
\email{yghuang@pku.edu.cn}

\author{Honggang Qi}
\affiliation{%
  \institution{University of Chinese Academy of Sciences}
  \city{Beijing}
  \country{China}
}
\email{hgqi@ucas.ac.cn}

\author{Geguang Pu}
\affiliation{%
 \institution{East China Normal University}
 \city{Shanghai}
 \country{China}}
\email{ggpu@sei.ecnu.edu.cn}

\author{Wanli Ouyang}
\affiliation{%
 \institution{Shanghai Artificial Intelligence Laboratory}
 \city{Shanghai}
 \country{China}}
\email{ouyangwanli@pjlab.org.cn}

\author{Xi Chen}
\authornote{Corresponding authors.} % 1. 在这里定义脚注文字，会自动生成一个符号（通常是 *）
\affiliation{%
  \institution{East China Normal University}
  \city{Shanghai}
  \country{China}}
\email{xchen@geo.ecnu.edu.cn}

\author{Jilin Hu}
\authornotemark[2] % 2. 这里使用 [1] 复用第一个 authornote 的符号，不再重复生成文字
\affiliation{%
  \institution{East China Normal University}
  \city{Shanghai}
  \country{China}}
\email{jlhu@dase.ecnu.edu.cn}


\begin{abstract}

While Time Series Foundation Models (TSFMs) have demonstrated exceptional performance in generalized forecasting, their performance often degrades significantly when deployed in real-world vertical domains characterized by temporal distribution shifts and domain-specific periodic structures. Current solutions are primarily constrained by two paradigms: Domain-Adaptive Pretraining (DAPT), which improves short-term domain fitting but frequently disrupts previously learned global temporal patterns due to catastrophic forgetting; and Retrieval-Augmented Generation (RAG), which incorporates external knowledge but introduces substantial retrieval overhead. This creates a severe scalability bottleneck that fails to meet the high-efficiency requirements of real-time stream processing. To break this impasse, we propose \textbf{MEM}ory for \textbf{T}ime \textbf{S}eries (MEMTS), a lightweight and plug-and-play method for retrieval-free domain adaptation in time series forecasting. The key component of MEMTS is a \textbf{K}nowledge \textbf{P}ersistence \textbf{M}odule (KPM), which internalizes domain-specific temporal dynamics, such as recurring seasonal patterns and trends into a compact set of learnable latent prototypes. In doing so, it transforms fragmented historical observations into continuous, parameterized knowledge representations. This paradigm shift enables MEMTS to achieve accurate domain adaptation with constant-time inference and near-zero latency, while effectively mitigating catastrophic forgetting of general temporal patterns, all without requiring any architectural modifications to the frozen TSFM backbone. Extensive experiments on multiple datasets demonstrate the SOTA performance of MEMTS.

\end{abstract}





\settopmatter{authorsperrow=4}
\maketitle

\section{Introduction}

\begin{figure}[t]
\centering
\includegraphics[width=0.48\textwidth]{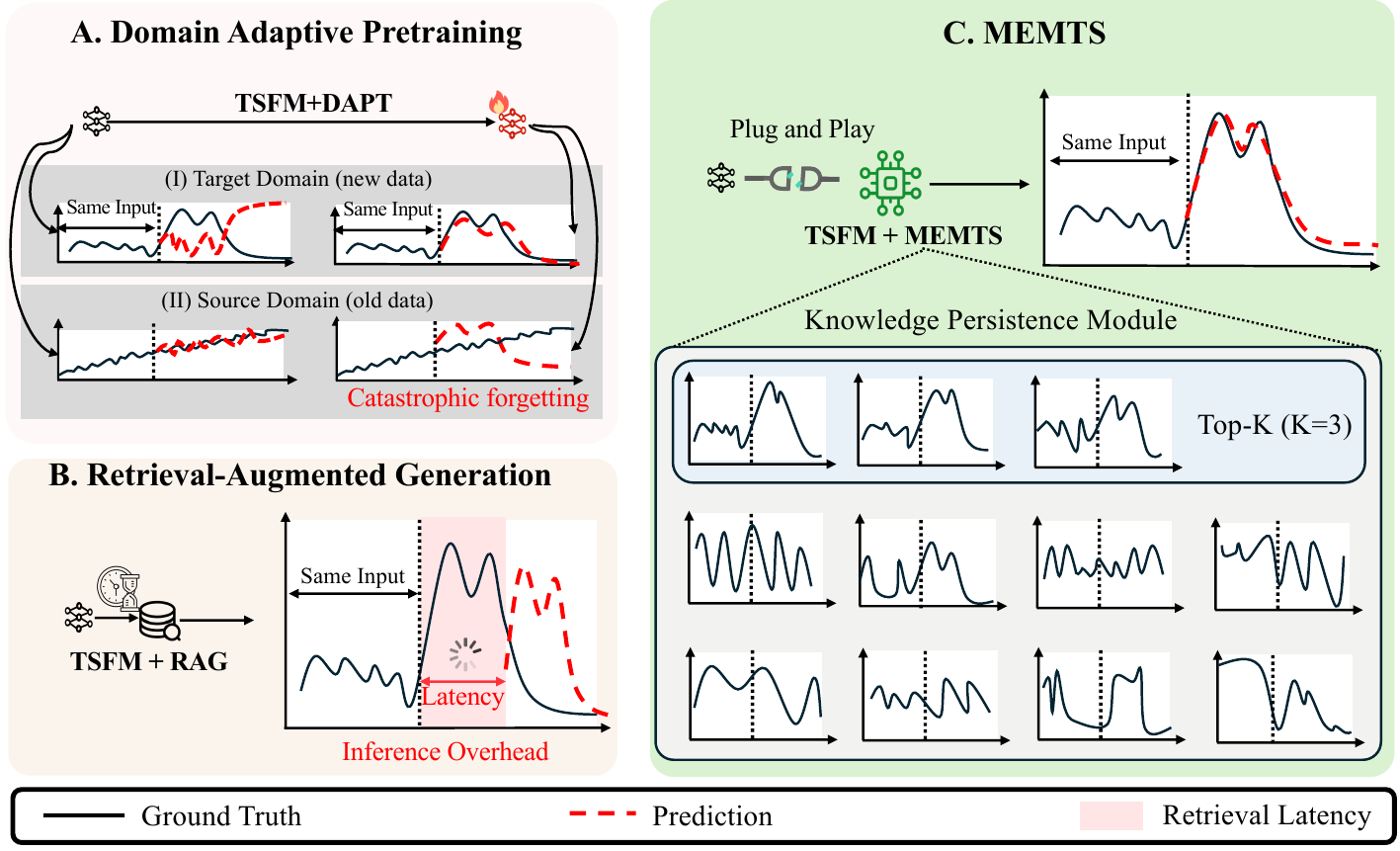} 
\caption{Comparison of domain adaptation paradigms. 
(A) Domain-Adaptive Pretraining (DAPT) fine-tunes the model, leading to accurate target adaptation but suffering from catastrophic forgetting on the source domain. 
(B) Retrieval-Augmented Generation (RAG) relies on external retrieval, introducing a significant inference latency gap before predictions can start.
(C) MEMTS (Ours) employs a plug-and-play Knowledge Persistence Module that internalizes patterns into parametric prototypes, enabling accurate adaptation with zero latency and no external storage.}
\vspace{-10pt}
\label{fig:introduction_compare}
\end{figure}

Time series forecasting leverages historical data to predict future values~\cite{qiu2025comprehensive,wu2025k2vae,qiu2024tfb,liu2025rethinking,qiu2025dag}, serving as a decision-making cornerstone in fields like finance \cite{qiu2025DBLoss,liu2026astgi,wu2025srsnet,qiu2025easytime,DBLP:conf/iiki/ZhangZDWW18} and transportation \cite{qiu2025duet,DBLP:conf/iclr/LiYS018,DBLP:journals/corr/abs-2004-08555}. Recently, the proliferation of large-scale and diverse datasets has driven a paradigm shift toward deep learning, owing to its strong ability to capture complex patterns in massive temporal data \cite{DBLP:journals/corr/abs-2004-13408, DBLP:conf/ijcai/WenZZCMY023, DBLP:conf/aaai/ZhouZPZLXZ21, DBLP:conf/nips/WuXWL21}. Building on this momentum, Time Series Foundation Models (TSFMs) have emerged as a powerful paradigm, shifting the focus from task-specific models to large-scale pretrained architectures (\emph{e.g.}, Lag-Llama \cite{rasul2023lag}, TimesFM \cite{DBLP:conf/icml/DasKSZ24}, MOIRAI \cite{DBLP:conf/icml/WooLKXSS24}).  

In contrast to Large Language Models (LLMs) that capture semantic commonalities in text \cite{DBLP:conf/nips/BrownMRSKDNSSAA20,DBLP:journals/corr/abs-2302-13971,DBLP:journals/corr/abs-2303-08774}, TSFMs aim to learn ``universal'' temporal patterns—such as trend, seasonality, and autocorr-elation—from massive cross-domain datasets \cite{DBLP:journals/corr/abs-2310-03589,DBLP:journals/tmlr/AnsariSTZMSSRPK24,DBLP:journals/corr/abs-2403-00131}. Specifically, a TSFM takes a historical ``look-back'' window of numerical sequences as input and generates a future ``forecast'' window as output. While the diversity of time series characteristics (\emph{e.g.}, differing trend or seasonal) makes the concept of a ``foundation'' more abstract than in Natural Language Processing (NLP), the success of TSFMs suggests that disparate domains do share underlying structural motifs \cite{DBLP:conf/icml/LiuZLH0L24,DBLP:journals/corr/abs-2303-06053}. However, when applying these time series foundation models to specific vertical domains, the mismatch between universally learned temporal patterns and domain-specific dynamics often results in performance degradation \cite{DBLP:conf/icml/JinPMWW22}. To bridge this gap, tailoring TSFMs to specialized domain through domain adaptation has become a critical necessity for practical deployment \cite{DBLP:journals/tist/WilsonC20,DBLP:conf/cikm/Du0FPQXW21}.

Despite its importance, efficient domain adaptation remains an open challenge. As illustrated in Figure \ref{fig:introduction_compare}, adapting TSFMs to specific domains generally follows two mainstream paradigms, each with inherent trade-offs. Domain-Adaptive Pretraining (DAPT) continues training TSFMs on domain-specific data to capture domain-specific temporal dynamics and unique patterns \cite{DBLP:conf/acl/GururanganMSLBD20,DBLP:conf/acl/RuderH18}, but it is computationally costly and often suffers from catastrophic forgetting, which degrades the model’s original zero-shot capabilities \cite{DBLP:journals/corr/KirkpatrickPRVD16,DBLP:journals/neco/FrenchC02}. Alternatively, Retrieval-Augmented Generation (RAG) avoids parameter updates by retrieving relevant historical segments from large datastores to enrich the input \cite{DBLP:conf/nips/LewisPPPKGKLYR020,DBLP:conf/icml/GuuLTPC20,DBLP:conf/icml/BorgeaudMHCRM0L22}. However, applying RAG to time series incurs substantial inference latency due to expensive nearest-neighbor searches (e.g., kNN \cite{DBLP:journals/tit/CoverH67}) and requires maintaining massive external vector repositories, which hinders its scalability in real-time forecasting scenarios \cite{DBLP:conf/iclr/KhandelwalLJZL20,DBLP:conf/eacl/IzacardG21}.

To better adapt TSFMs to specific domains, we propose a plug-and-play parametric memory method, called \textbf{MEMTS} (\textbf{MEM}ory for \textbf{T}ime \textbf{S}eries), to enable efficient domain adaptation of TSFMs without modifying their original parameters. At its core, MEMTS introduces a \textbf{K}nowledge \textbf{P}ersistence \textbf{M}odule (KPM) that internalizes domain-specific temporal dynamics into a compact set of learnable latent prototypes. Unlike RAG, which relies on explicit retrieval from external archives, MEMTS operates as a purely parametric mechanism.  The KPM comprises a Context Encoder that captures historical motifs and a Future-Separable Decoder that generates diverse future hypotheses, optimized with a set-matching permutation loss to prevent mode collapse. We further propose an Adaptive Fusion module that fuses memory-derived candidates with the frozen TSFM predictions and dynamically modulates the contribution of domain knowledge to produce precise, time-step-wise corrections. Owing to its architectural independence, MEMTS can be seamlessly attached to any backbone TSFM (e.g., Sundial \cite{liu2025sundial}, Chronos \cite{DBLP:journals/tmlr/AnsariSTZMSSRPK24}) once trained on a target domain, delivering domain-specific enhancements in a ``one-to-many'' manner while substantially reducing computational and deployment overhead. Ultimately, MEMTS formalizes a high-efficiency, low-latency paradigm for domain adaptation in the time series landscape.

Our contributions are summarized as follows:

\begin{itemize}[leftmargin=*, topsep=1pt, partopsep=0pt, itemsep=2pt, parsep=0pt]
    
    \item To better adapt TSFMs to specific domains, we propose \textbf{MEMTS}, a plug-and-play parametric memory method that enables efficient domain adaptation without updating the original TSFM parameters.

    \item We design a Knowledge Persistence Module (KPM) that internalizes domain-specific temporal dynamics into learnable latent prototypes, emulating retrieval behavior while enabling near-zero-latency knowledge injection.
    
    \item We conduct extensive experiments on multiple datasets. The results demonstrate the effectiveness of \textbf{MEMTS}, consistently outperforming SOTA baselines with low deployment overhead.
\end{itemize}

\section{Related Work}

\subsection{Time Series Foundation Models}

Early methods for time series forecasting were mainly based on statistical modeling and signal processing theories, such as ARIMA. These approaches offer good interpretability in modeling trends and seasonality.With the development of deep learning, researchers have proposed models based on RNNs \cite{elman1990finding}, LSTMs \cite{graves2012long}, Temporal CNNs \cite{lea2017temporal}, and attention mechanisms, which significantly outperform traditional methods in modeling complex patterns \cite{DBLP:journals/bigdata/TorresHSMT21} . However, these models usually need to be trained on specific tasks and datasets, and their generalization ability is limited in cross-domain or distribution-shift scenarios.

In recent years, inspired by the success of foundation models in natural language processing and computer vision, large-scale pre-trained models have gradually emerged in the time series domain, known as Time Series Foundation Models (TSFMs) \cite{DBLP:conf/icml/WooLKXSS24} . Representative works include Sundial \cite{liu2025sundial}, Moirai \cite{DBLP:conf/icml/WooLKXSS24} , and Chronos \cite{DBLP:journals/tmlr/AnsariSTZMSSRPK24}, among others.

\begin{figure*}[t]

\centering
\includegraphics[height=0.35\textwidth]{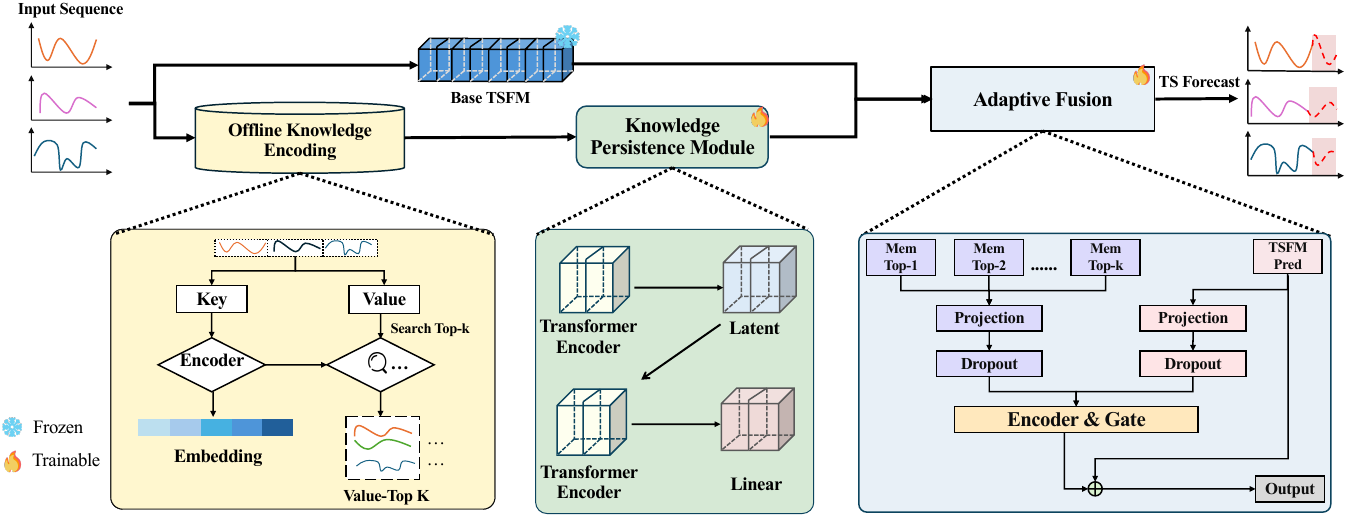}
\caption{Overview of MEMTS: A parametric memory knowledge enhancer for time series foundation models. The framework consists of three key components: (1) Knowledge Building module that extracts and stores domain-specific patterns, (2) Knowledge Persistence Module that generates memory-enhanced representations, and (3) Adaptive Fusion module that combines base model predictions with memory-derived knowledge to produce enhanced forecasts.}
\label{fig:MEMTS-arch2}
\end{figure*}

\subsection{Domain Adaptation and Knowledge Enhancement}

However, adapting them to specific domains to achieve optimal performance remains a significant challenge. While full-parameter fine-tuning for domain adaptation can be effective in some cases, it is typically computationally expensive and prone to catastrophic forgetting \cite{DBLP:conf/icml/ZhangLZY24}.

Recently, several studies have attempted to introduce the RAG paradigm into time series forecasting. For example, by retrieving and concatenating similar historical segments, these methods enhance the zero-shot forecasting capability of time series foundation models \cite{ning2025ts}. However, such approaches still have limitations: on the one hand, similarity search over high-dimensional sequences introduces significant inference latency; on the other hand, the knowledge obtained through retrieval is only used as external context and is not internalized into generalizable model parameters. Another line of research that has developed in parallel with retrieval-based methods is memory mechanisms \cite{graves2012long,DBLP:conf/cikm/LiuLZGC22}. 

While recent work has explored parameterized memory in Natural Language Processing (NLP), such as the Memory Decoder \cite{cao2025memory}, these methods are primarily optimized for discrete token distributions. In contrast, time series data are characterized by high continuity and intense volatility, and inherently require the modeling of multi-modal future trajectories. MEMTS explicitly addresses these domain-specific challenges by introducing the Knowledge Persistence Module (KPM) and a dedicated permutation loss, both of which are specifically designed to internalize complex structural motifs and capture features that general-purpose NLP-inspired memory paradigms fail to model.

% Related work introduces learnable external memory or long-term state modules, enabling models to explicitly store and recall historical information, thereby alleviating problems of long-term dependencies and forgetting \cite{trinh2018learning,santoro2016meta,mezghan2022memory}. However, these models are often tightly coupled with specific network architectures and require end-to-end joint training, making them difficult to apply directly to frozen foundation models and rarely providing explicit modeling for domain-specific patterns.

Overall, MEMTS addresses a critical gap commonly faced by time series foundation models in real-world deployment: how to inject key domain-specific historical patterns into the model in a controllable, interpretable, and generalizable manner without modifying the parameters of the host model and without requiring online retrieval. 

\section{Preliminaries}

\subsection{Definitions}
For a given time series $\mathbf{x} = (x_1, x_2, \ldots, x_T) \in \mathbb{R}^{T \times D}$, we define the historical \textit{look-back window} (key) at time $t$ as:
\[
\mathbf{k}_t = [x_t, x_{t+1}, \ldots, x_{t+K-1}] \in \mathbb{R}^{K \times D},
\]
and its corresponding future \textit{forecast window} (value) as:
\[
\mathbf{v}_t = [x_{t+K}, x_{t+K+1}, \ldots, x_{t+K+V-1}] \in \mathbb{R}^{V \times D}.
\]

\subsection{Problem Statement}
Given a frozen Time Series Foundation Model (TSFM) $\mathcal{M}_\text{base}$ pretrained on large-scale data, and a target-domain dataset $\mathcal{X}_\text{target}$, our goal is to improve the zero-shot forecasting performance of $\mathcal{M}_\text{base}$ on $\mathcal{X}_\text{target}$ without updating its parameters. We achieve this by learning a lightweight, plug-and-play parameterized memory module $\mathcal{M}_\text{mem}$ that captures domain-specific temporal patterns and seamlessly integrates with $\mathcal{M}_\text{base}$ during inference.

\section{Methodology}
\textbf{Overview:} As illustrated in Figure \ref{fig:MEMTS-arch2}, we propose a comprehensive forecasting pipeline that integrates our novel MEMTS framework with existing Time Series Foundation Models (TSFMs). The workflow begins with an input sequence processed simultaneously via two parallel pathways: the input is fed into a frozen Base TSFM to obtain a preliminary forecast, while concurrently being processed by our Offline Knowledge Encoding and KPM (Section \ref{sec:kb_construction} \& \ref{sec:kpm}) to retrieve and generate domain-specific future candidates. Finally, these two distinct information sources are dynamically merged by the Adaptive Fusion module (Section \ref{sec:adaptive_fusion}) to yield the final enhanced output forecast.

\subsection{Offline Knowledge Internalization}
\label{sec:kb_construction}
To mitigate the online retrieval bottleneck, MEMTS internalizes domain-specific patterns into a set of \textit{parameterized prototypes}. As illustrated in Fig.~\ref{fig:MEMTS-arch2}, this phase follows a three-stage pipeline: (i) temporal segmentation, (ii) latent projection, and (iii) leakage-free selection.

\paragraph{Temporal Segmentation and Partitioning.}
Given a time series collection $\mathcal{X}$, we employ a sliding-window strategy to partition sequences into key-value pairs $(\mathbf{k}^{(i)}_t, \mathbf{v}^{(i)}_t)$ \cite{SALINAS20201181}:
\begin{equation}
\mathbf{k}^{(i)}_t = [x^{(i)}_t, \dots, x^{(i)}_{t+K-1}], \quad \mathbf{v}^{(i)}_t = [x^{(i)}_{t+K}, \dots, x^{(i)}_{t+K+V-1}],
\end{equation}
where $K$ and $V$ denote the context length and prediction horizon. To ensure evaluation integrity, we define temporal split points $t^{(i)}_{\text{train}} = \lfloor \alpha L_i \rfloor$ and $t^{(i)}_{\text{val}} = \lfloor (\alpha + \beta) L_i \rfloor$ for each sequence of length $L_i$. A pair is assigned to a split only if its entire span $[t, t+K+V-1]$ lies fully within the respective temporal interval \cite{hyndman2018forecasting}, strictly preventing cross-split leakage.

\paragraph{Latent Knowledge Projection and Indexing.}
Each historical segment $\mathbf{k}^{(i)}_t$ is mapped into a $d$-dimensional latent space:
\begin{equation}
\mathbf{z}^{(i)}_t = f_{\text{enc}}(\mathbf{k}^{(i)}_t) \in \mathbb{R}^{d},
\end{equation}
where $f_{\text{enc}}(\cdot)$ captures local structural motifs. The resulting memory units $(\mathbf{z}^{(i)}_t, \mathbf{v}^{(i)}_t)$ are organized using an inverted file (IVF) structure to enable efficient similarity search during the training phase.

\paragraph{Leakage-Free Multi-Future Selection.}
To capture stochasticity, we identify the top-$k$ most similar encoded keys from memory. To prevent the model from exploiting temporal proximity, a candidate $(i_n, t_n)$ is excluded if it overlaps with the query $(i_q, t_q)$:
\begin{equation}
i_n = i_q \quad \text{and} \quad |t_n - t_q| < K+V.
\end{equation}
This rigorous masking ensures that the KPM learns robust mappings to multiple plausible futures without data leakage.

\begin{figure}[t]
\centering
\includegraphics[width=0.48\textwidth]{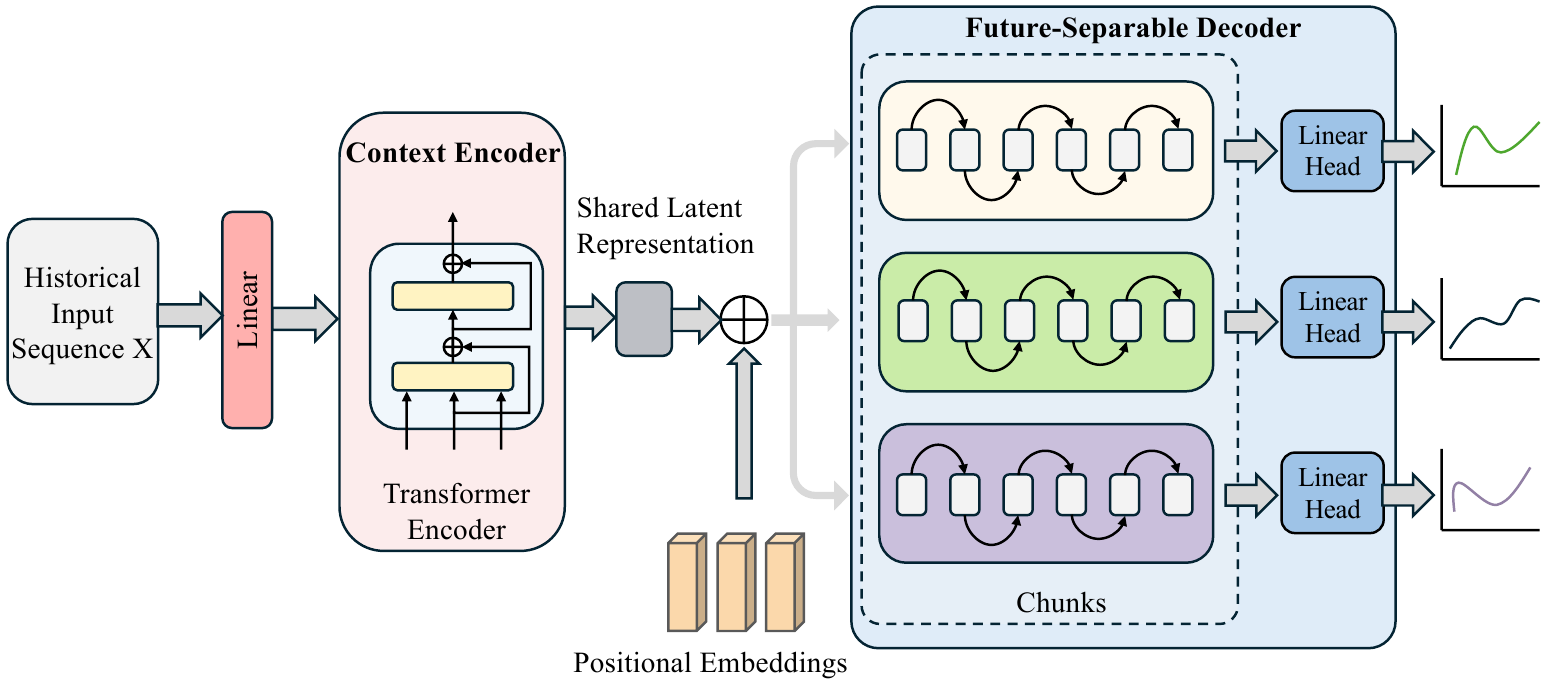}
\caption{Architecture of the KPM model for time-series forecasting. Historical inputs are encoded into a shared latent representation, which is fed into a Future-Separable Decoder with parallel decoding chunks to generate multiple independent future sequences.}
\vspace{-1em}
\label{fig:kpm架构图}
\end{figure}

\subsection{Knowledge Persistence Module (KPM)}
\label{sec:kpm}

The KPM is designed as a set-output, future-separable decoder that transforms latent historical embeddings into multiple plausible evolution hypotheses. As a decoupled enhancement, KPM enables zero-latency knowledge injection while keeping the base TSFM frozen.
% [New]: 新增过渡句，自然引出 Figure 3，作为 Figure 2 的补充
While Figure \ref{fig:MEMTS-arch2} presents the high-level position of KPM within the MEMTS framework, Figure \ref{fig:kpm架构图} provides a detailed internal view of its architecture, which consists of two primary blocks: a Context Encoder and a Future-Separable Decoder.

\paragraph{Stage 1: Contextual Latent Encoding.}
Given the historical embedding $\mathbf{z} \in \mathbb{R}^{d}$, we first apply a linear projection and a Transformer-based encoder $f_{\text{enc}}(\cdot)$ to map it into a shared latent space.
% [Modified]: 增加关于kpm架构图的描述
As depicted in the Context Encoder block (left panel of Figure \ref{fig:kpm架构图}), this process yields:
\begin{equation}
\mathbf{h} = f_{\text{enc}}(\mathbf{z}) \in \mathbb{R}^{d_h},
\end{equation}
where $\mathbf{h}$ serves as a universal conditioning variable. This stage internalizes global temporal motifs, providing a persistent reference base that eliminates the need for expensive online database queries.

\paragraph{Stage 2: Multi-Branch Parallel Decoding.}
To capture the multi-modal nature of future evolutions, we partition the horizon $V$ into $T$ segments of length $c$ ($V = T \cdot c$). For $M$ parallel branches, the $t$-th segment representation $\mathbf{q}_{m,t}$ is initialized by broadcasting the shared context $\mathbf{h}$ and integrating it with the Positional Embeddings shown at the bottom of Figure \ref{fig:kpm架构图}: % 新增架构图描述
\begin{equation}
\mathbf{q}_{m,t} = \mathbf{h} + \mathbf{e}^{(m)}_{\text{query},t} + \mathbf{e}^{(m)}_{\text{pos},t},
\end{equation}
where $\mathbf{e}^{(m)}_{\text{query},t}$ and $\mathbf{e}^{(m)}_{\text{pos},t}$ are branch-specific learnable queries and positional embeddings. 

These tokens are then fed into the Future-Separable Decoder (right panel of Figure \ref{fig:kpm架构图}). To ensure inter-branch independence (i.e., no information leakage between future hypotheses), intra-branch dependencies are modeled via a shared-weight Transformer encoder $\mathcal{F}(\cdot)$ applied independently to each branch:
\begin{equation}
\tilde{\mathbf{Q}}_{m} = \mathcal{F}(\mathbf{Q}_{m}), \quad \text{with } \mathbf{Q}_{m} = [\mathbf{q}_{m,1}, \dots, \mathbf{q}_{m,T}]^\top \in \mathbb{R}^{T \times d_h}.
\end{equation}

Finally, each latent segment is mapped to the numerical space via a linear head $g(\cdot)$ (visualized as "Linear Head" in Figure \ref{fig:kpm架构图}) and concatenated to form the branch forecast:
\begin{equation}
\hat{\mathbf{y}}_{m} = \mathrm{Concat}(g(\tilde{\mathbf{q}}_{m,1}), \dots, g(\tilde{\mathbf{q}}_{m,T})) \in \mathbb{R}^{V}.
\end{equation}

\paragraph{Permutation-Invariant Optimization.}
To avoid "mean collapse"—where the model converges to a blurry average of possible futures—we employ a set-matching loss \cite{10.1145/3502728} during the offline phase:
\begin{equation}
\mathcal{L}_{\text{mem}} = \min_{\pi \in \Pi_M} \sum_{m=1}^{M} \left\| \hat{\mathbf{y}}_{m} - \mathbf{y}_{\pi(m)} \right\|_2^2,
\end{equation}
where $\Pi_M$ denotes all permutations of $M$ elements, and $\mathbf{y}_{\pi(m)}$ are targets from the knowledge base. This encourages each branch to specialize in distinct future patterns, enabling KPM to provide precise corrections during abrupt distribution shifts.

\subsection{Adaptive Fusion}
\label{sec:adaptive_fusion}

To integrate domain-specific motifs from KPM with the frozen Base TSFM, we propose the Adaptive Fusion (AF) module. AF functions as a gated residual adapter that refines the TSFM's generalist predictions without altering its original parameters, thereby preserving zero-shot capabilities.

\paragraph{Feature Alignment and Tokenization.}
We first bridge the representational gap between the frozen foundation model and the learnable memory. Given the base prediction $\mathbf{b} \in \mathbb{R}^V$ and memory candidates $\{\hat{\mathbf{y}}_m\}_{m=1}^M$, we project them into a $d$-dimensional latent space:
\begin{equation}
\mathbf{h}_m = \mathrm{Drop}(\phi_{\text{mem}}(\hat{\mathbf{y}}_m)), \quad \mathbf{h}_b = \mathrm{Drop}(\phi_{\text{base}}(\mathbf{b})),
\end{equation}
where $\phi(\cdot)$ denotes bottleneck projection layers. We employ an asymmetric dropout strategy with a higher rate $p_{\text{mem}}$ for memory tokens to enhance robustness against noisy prototypes. Learnable positional embeddings are added to these tokens to preserve functional distinctions.

\paragraph{Cross-View Contextual Pooling.}
To model inter-dependencies between the generalist (TSFM) and specialist (KPM) perspectives, aligned tokens are processed by a Transformer encoder to capture cross-branch correlations. Let $\{\mathbf{z}_i\}_{i=1}^{M+1}$ be the encoder outputs. We introduce a dynamic weighting mechanism to pool this knowledge:
\begin{equation}
w_i = \frac{\exp(g(\mathbf{z}_i) / \tau)}{\sum_{j=1}^{M+1} \exp(g(\mathbf{z}_j) / \tau)}, \quad \mathbf{z}_{\text{fused}} = \sum_{i=1}^{M+1} w_i \mathbf{z}_i,
\end{equation}
where $g(\cdot)$ is a learnable scoring function and $\tau$ is a temperature coefficient. This allows the model to selectively prioritize memory branches that best align with the current temporal manifold.

\paragraph{Gated Residual Correction.}
The final forecast $\hat{\mathbf{y}}$ is generated via a gated residual connection:
\begin{equation}
\hat{\mathbf{y}} = \mathbf{b} + \mathbf{g} \odot \tanh(f_{\text{head}}(\mathbf{z}_{\text{fused}})),
\end{equation}
where $f_{\text{head}}(\cdot)$ is the prediction head and $\mathbf{g} \in \mathbb{R}^V$ is a learnable stepwise gate. This structure provides a strong inductive bias toward the foundation model while enabling fine-grained, time-step-wise corrections based on internalized domain knowledge.

\begin{table*}[htbp]
  \centering
  \caption{Zero-shot forecasting comparison between foundation models and MEMTS across 6 datasets. Horizons $H \in \{96, 192, 336, 720\}$. Bolt-S/B denotes Chronos-Bolt-Small/Base. The last row of each dataset (highlighted in gray) shows the average performance.}
  \resizebox{\textwidth}{!}{
    \begin{tabular}{c|c|cccc|cccc|cccc|cccc|cccc}
    \toprule
    \multicolumn{2}{c|}{\multirow{2}[4]{*}{Models}} & \multicolumn{4}{c|}{\textbf{Bolt-S}} & \multicolumn{4}{c|}{\textbf{Bolt-B}} & \multicolumn{4}{c|}{\textbf{Moment}}     & \multicolumn{4}{c|}{\textbf{Moirai}}    & \multicolumn{4}{c}{\textbf{Sundial}} \\
\cmidrule{3-22}    \multicolumn{2}{c|}{} & \multicolumn{2}{c}{Ori} & \multicolumn{2}{c|}{+MEMTS} & \multicolumn{2}{c}{Ori} & \multicolumn{2}{c|}{+MEMTS} & \multicolumn{2}{c}{Ori} & \multicolumn{2}{c|}{+MEMTS} & \multicolumn{2}{c}{Ori} & \multicolumn{2}{c|}{+MEMTS} & \multicolumn{2}{c}{Ori} & \multicolumn{2}{c}{+MEMTS} \\
    \midrule
    \multicolumn{2}{c|}{Metric} & MSE    & MAE    & MSE    & MAE    & MSE    & MAE    & MSE    & MAE    & MSE    & MAE    & MSE    & MAE    & MSE    & MAE    & MSE    & MAE    & MSE    & MAE    & MSE    & MAE \\
    \midrule
    \multirow{5}[2]{*}{ECL} & 96    & 0.360  & \textbf{0.411 } & \textbf{0.327 } & 0.418  & 0.329  & 0.389  & \textbf{0.266 } & \textbf{0.354 } & 0.867  & 0.767  & \textbf{0.860 } & \textbf{0.765 } & 1.412  & 0.906  & \textbf{0.644 } & \textbf{0.624 } & 0.207  & \textbf{0.286 } & \textbf{0.204 } & 0.287  \\
          & 192   & 0.242  & 0.333  & \textbf{0.184 } & \textbf{0.286 } & 0.241  & 0.333  & \textbf{0.185 } & \textbf{0.295 } & 0.877  & \textbf{0.772 } & \textbf{0.874 } & \textbf{0.772 } & \textbf{0.191 } & \textbf{0.277 } & 0.193  & 0.288  & 0.167  & \textbf{0.260 } & \textbf{0.166 } & 0.261  \\
          & 336   & 0.238  & 0.327  & \textbf{0.202 } & \textbf{0.303 } & 0.236  & 0.325  & \textbf{0.200 } & \textbf{0.310 } & 0.891  & 0.777  & \textbf{0.885 } & \textbf{0.775 } & 1.244  & 0.867  & \textbf{0.848 } & \textbf{0.726 } & 0.185  & \textbf{0.278 } & \textbf{0.183 } & \textbf{0.278 } \\
          & 720   & 0.264  & 0.340  & \textbf{0.242 } & \textbf{0.328 } & 0.253  & 0.331  & \textbf{0.235 } & \textbf{0.326 } & 0.905  & 0.779  & \textbf{0.900 } & \textbf{0.778 } & 1.233  & 0.869  & \textbf{0.998 } & \textbf{0.785 } & 0.220  & 0.308  & \textbf{0.217 } & \textbf{0.307 } \\
          & \cellcolor{gray!15}Avg   & \cellcolor{gray!15}0.276  & \cellcolor{gray!15}0.353  & \cellcolor{gray!15}\textbf{0.239 } & \cellcolor{gray!15}\textbf{0.334 } & \cellcolor{gray!15}0.265  & \cellcolor{gray!15}0.345  & \cellcolor{gray!15}\textbf{0.222 } & \cellcolor{gray!15}\textbf{0.321 } & \cellcolor{gray!15}0.885  & \cellcolor{gray!15}0.774  & \cellcolor{gray!15}\textbf{0.880 } & \cellcolor{gray!15}\textbf{0.773 } & \cellcolor{gray!15}1.020  & \cellcolor{gray!15}0.730  & \cellcolor{gray!15}\textbf{0.671 } & \cellcolor{gray!15}\textbf{0.606 } & \cellcolor{gray!15}0.195  & \cellcolor{gray!15}0.283  & \cellcolor{gray!15}\textbf{0.193 } & \cellcolor{gray!15}\textbf{0.283 } \\
    \midrule
    \multirow{5}[2]{*}{AusRain} & 96    & 2.225  & 1.198  & \textbf{1.337 } & \textbf{0.914 } & 2.077  & 1.150  & \textbf{1.321 } & \textbf{0.905 } & 1.170  & 0.880  & \textbf{1.153 } & \textbf{0.874 } & 2.261  & 1.176  & \textbf{1.506 } & \textbf{0.981 } & 1.240  & 0.878  & \textbf{1.199 } & \textbf{0.865 } \\
          & 192   & 1.653  & 1.036  & \textbf{1.102 } & \textbf{0.843 } & 1.676  & 1.041  & \textbf{1.143 } & \textbf{0.856 } & 1.086  & 0.853  & \textbf{1.078 } & \textbf{0.850 } & 1.123  & 0.849  & \textbf{1.098 } & \textbf{0.841 } & 1.213  & 0.880  & \textbf{1.178 } & \textbf{0.868 } \\
          & 336   & 1.421  & 0.960  & \textbf{1.116 } & \textbf{0.853 } & 1.461  & 0.972  & \textbf{1.132 } & \textbf{0.858 } & 1.075  & 0.849  & \textbf{1.068 } & \textbf{0.847 } & 1.950  & 1.095  & \textbf{1.587 } & \textbf{1.009 } & 1.226  & 0.890  & \textbf{1.199 } & \textbf{0.881 } \\
          & 720   & 1.309  & 0.923  & \textbf{1.186 } & \textbf{0.881 } & 1.578  & 1.014  & \textbf{1.405 } & \textbf{0.955 } & 1.073  & 0.848  & \textbf{1.067 } & \textbf{0.846 } & 1.756  & 1.055  & \textbf{1.643 } & \textbf{1.030 } & 1.230  & 0.897  & \textbf{1.215 } & \textbf{0.892 } \\
          & \cellcolor{gray!15}Avg   & \cellcolor{gray!15}1.652  & \cellcolor{gray!15}1.029  & \cellcolor{gray!15}\textbf{1.185 } & \cellcolor{gray!15}\textbf{0.873 } & \cellcolor{gray!15}1.698  & \cellcolor{gray!15}1.044  & \cellcolor{gray!15}\textbf{1.250 } & \cellcolor{gray!15}\textbf{0.894 } & \cellcolor{gray!15}1.101  & \cellcolor{gray!15}0.858  & \cellcolor{gray!15}\textbf{1.092 } & \cellcolor{gray!15}\textbf{0.854 } & \cellcolor{gray!15}1.773  & \cellcolor{gray!15}1.044  & \cellcolor{gray!15}\textbf{1.459 } & \cellcolor{gray!15}\textbf{0.965 } & \cellcolor{gray!15}1.227  & \cellcolor{gray!15}0.887  & \cellcolor{gray!15}\textbf{1.197 } & \cellcolor{gray!15}\textbf{0.877 } \\
    \midrule
    \multirow{5}[2]{*}{METR-LA} & 96    & 2.406  & 1.092  & \textbf{1.440 } & \textbf{0.779 } & 2.436  & 1.089  & \textbf{1.750 } & \textbf{0.873 } & 1.386  & 0.765  & \textbf{1.346 } & \textbf{0.762 } & 1.570  & 0.758  & \textbf{1.217 } & \textbf{0.666 } & 1.368  & 0.684  & \textbf{1.327 } & \textbf{0.674 } \\
          & 192   & 1.905  & 0.928  & \textbf{1.398 } & \textbf{0.797 } & 2.018  & 0.958  & \textbf{1.514 } & \textbf{0.798 } & 1.281  & 0.789  & \textbf{1.261 } & \textbf{0.787 } & 1.484  & \textbf{0.645 } & \textbf{1.440 } & 0.649  & 1.336  & 0.654  & \textbf{1.309 } & \textbf{0.652 } \\
          & 336   & 1.836  & 0.873  & \textbf{1.502 } & \textbf{0.829 } & 1.916  & 0.889  & \textbf{1.565 } & \textbf{0.821 } & 1.384  & 0.820  & \textbf{1.359 } & \textbf{0.817 } & 1.754  & 0.865  & \textbf{1.427 } & \textbf{0.790 } & 1.412  & 0.695  & \textbf{1.389 } & \textbf{0.694 } \\
          & 720   & 2.223  & 0.972  & \textbf{2.035 } & \textbf{0.936 } & 2.308  & 0.982  & \textbf{2.072 } & \textbf{0.964 } & 1.671  & 0.930  & \textbf{1.637 } & \textbf{0.922 } & 2.055  & 0.974  & \textbf{1.823 } & \textbf{0.916 } & 1.628  & 0.792  & \textbf{1.608 } & \textbf{0.790 } \\
          & \cellcolor{gray!15}Avg   & \cellcolor{gray!15}2.092  & \cellcolor{gray!15}0.966  & \cellcolor{gray!15}\textbf{1.594 } & \cellcolor{gray!15}\textbf{0.835 } & \cellcolor{gray!15}2.170  & \cellcolor{gray!15}0.979  & \cellcolor{gray!15}\textbf{1.725 } & \cellcolor{gray!15}\textbf{0.864 } & \cellcolor{gray!15}1.430  & \cellcolor{gray!15}0.826  & \cellcolor{gray!15}\textbf{1.401 } & \cellcolor{gray!15}\textbf{0.822 } & \cellcolor{gray!15}1.716  & \cellcolor{gray!15}0.810  & \cellcolor{gray!15}\textbf{1.477 } & \cellcolor{gray!15}\textbf{0.755 } & \cellcolor{gray!15}1.436  & \cellcolor{gray!15}0.706  & \cellcolor{gray!15}\textbf{1.408 } & \cellcolor{gray!15}\textbf{0.702 } \\
    \midrule
    \multirow{5}[2]{*}{PEMS04} & 96    & 2.214  & 1.188  & \textbf{0.775 } & \textbf{0.677 } & 2.189  & 1.177  & \textbf{1.156 } & \textbf{0.836 } & 2.045  & 1.238  & \textbf{1.961 } & \textbf{1.213 } & 2.092  & 1.188  & \textbf{0.822 } & \textbf{0.731 } & 1.481  & 0.943  & \textbf{1.385 } & \textbf{0.910 } \\
          & 192   & 0.926  & 0.718  & \textbf{0.701 } & \textbf{0.657 } & 1.127  & 0.794  & \textbf{0.795 } & \textbf{0.697 } & 1.132  & 0.918  & \textbf{1.123 } & \textbf{0.915 } & 0.260  & \textbf{0.308 } & \textbf{0.255 } & 0.319  & \textbf{0.191 } & \textbf{0.273 } & \textbf{0.191 } & 0.278  \\
          & 336   & 0.840  & 0.675  & \textbf{0.774 } & \textbf{0.683 } & 0.975  & 0.725  & \textbf{0.880 } & \textbf{0.717 } & 1.057  & 0.882  & \textbf{1.052 } & \textbf{0.881 } & 1.737  & 1.043  & \textbf{1.372 } & \textbf{0.940 } & \textbf{0.214 } & \textbf{0.292 } & \textbf{0.214 } & 0.295  \\
          & 720   & 0.574  & \textbf{0.527 } & \textbf{0.559 } & 0.534  & 0.616  & \textbf{0.538 } & \textbf{0.614 } & 0.565  & 1.101  & 0.902  & \textbf{1.094 } & \textbf{0.900 } & 1.736  & 1.052  & \textbf{1.567 } & \textbf{1.015 } & \textbf{0.220 } & \textbf{0.305 } & 0.221  & 0.307  \\
          & \cellcolor{gray!15}Avg   & \cellcolor{gray!15}1.139  & \cellcolor{gray!15}0.777  & \cellcolor{gray!15}\textbf{0.702 } & \cellcolor{gray!15}\textbf{0.638 } & \cellcolor{gray!15}1.227  & \cellcolor{gray!15}0.809  & \cellcolor{gray!15}\textbf{0.861 } & \cellcolor{gray!15}\textbf{0.704 } & \cellcolor{gray!15}1.334  & \cellcolor{gray!15}0.985  & \cellcolor{gray!15}\textbf{1.308 } & \cellcolor{gray!15}\textbf{0.977 } & \cellcolor{gray!15}1.456  & \cellcolor{gray!15}0.898  & \cellcolor{gray!15}\textbf{1.004 } & \cellcolor{gray!15}\textbf{0.751 } & \cellcolor{gray!15}0.527  & \cellcolor{gray!15}0.453  & \cellcolor{gray!15}\textbf{0.502 } & \cellcolor{gray!15}\textbf{0.448 } \\
    \midrule
    \multirow{5}[2]{*}{PEMS08} & 96    & 2.204  & 1.189  & \textbf{0.827 } & \textbf{0.715 } & 2.186  & 1.185  & \textbf{1.191 } & \textbf{0.867 } & 2.075  & 1.243  & \textbf{1.989 } & \textbf{1.218 } & 1.881  & 1.094  & \textbf{0.837 } & \textbf{0.735 } & 1.403  & 0.910  & \textbf{1.314 } & \textbf{0.879 } \\
          & 192   & 1.028  & 0.768  & \textbf{0.774 } & \textbf{0.696 } & 1.192  & 0.840  & \textbf{0.877 } & \textbf{0.735 } & 1.147  & 0.912  & \textbf{1.137 } & \textbf{0.909 } & 0.232  & \textbf{0.288 } & \textbf{0.228 } & 0.298  & \textbf{0.167 } & \textbf{0.255 } & \textbf{0.167 } & 0.261  \\
          & 336   & 0.987  & 0.743  & \textbf{0.864 } & \textbf{0.729 } & 1.068  & 0.781  & \textbf{0.976 } & \textbf{0.765 } & 1.067  & 0.876  & \textbf{1.061 } & \textbf{0.875 } & 1.719  & 1.023  & \textbf{1.355 } & \textbf{0.925 } & 0.183  & \textbf{0.271 } & \textbf{0.182 } & 0.274  \\
          & 720   & 0.620  & 0.562  & \textbf{0.593 } & \textbf{0.559 } & 0.684  & \textbf{0.589 } & \textbf{0.669 } & 0.604  & 1.088  & 0.888  & \textbf{1.081 } & \textbf{0.886 } & 1.712  & 1.030  & \textbf{1.554 } & \textbf{0.998 } & \textbf{0.191 } & \textbf{0.282 } & \textbf{0.191 } & 0.285  \\
          & \cellcolor{gray!15}Avg   & \cellcolor{gray!15}1.210  & \cellcolor{gray!15}0.815  & \cellcolor{gray!15}\textbf{0.764 } & \cellcolor{gray!15}\textbf{0.675 } & \cellcolor{gray!15}1.283  & \cellcolor{gray!15}0.849  & \cellcolor{gray!15}\textbf{0.928 } & \cellcolor{gray!15}\textbf{0.743 } & \cellcolor{gray!15}1.344  & \cellcolor{gray!15}0.980  & \cellcolor{gray!15}\textbf{1.317 } & \cellcolor{gray!15}\textbf{0.972 } & \cellcolor{gray!15}1.386  & \cellcolor{gray!15}0.859  & \cellcolor{gray!15}\textbf{0.994 } & \cellcolor{gray!15}\textbf{0.739 } & \cellcolor{gray!15}0.486  & \cellcolor{gray!15}0.430  & \cellcolor{gray!15}\textbf{0.464 } & \cellcolor{gray!15}\textbf{0.425 } \\
    \midrule
    \multirow{5}[2]{*}{Solar} & 96    & 1.257  & 0.699  & \textbf{0.472 } & \textbf{0.462 } & 1.277  & 0.685  & \textbf{0.715 } & \textbf{0.547 } & 1.071  & 0.782  & \textbf{1.044 } & \textbf{0.771 } & 1.440  & 0.929  & \textbf{0.553 } & \textbf{0.558 } & 0.907  & 0.594  & \textbf{0.838 } & \textbf{0.574 } \\
          & 192   & 0.931  & \textbf{0.529 } & \textbf{0.668 } & 0.577  & 0.893  & \textbf{0.497 } & \textbf{0.646 } & 0.596  & 0.848  & \textbf{0.728 } & \textbf{0.842 } & 0.731  & 0.915  & \textbf{0.522 } & \textbf{0.830 } & 0.525  & 0.372  & \textbf{0.314 } & \textbf{0.361 } & 0.321  \\
          & 336   & 0.980  & \textbf{0.566 } & \textbf{0.854 } & 0.633  & 0.948  & \textbf{0.534 } & \textbf{0.794 } & 0.611  & 0.850  & \textbf{0.732 } & \textbf{0.844 } & 0.734  & 1.183  & 0.907  & \textbf{1.013 } & \textbf{0.813 } & 0.390  & \textbf{0.330 } & \textbf{0.382 } & 0.336  \\
          & 720   & 0.812  & \textbf{0.487 } & \textbf{0.731 } & 0.534  & 0.847  & \textbf{0.493 } & \textbf{0.767 } & 0.571  & 0.824  & \textbf{0.729 } & \textbf{0.819 } & 0.731  & 1.211  & 0.928  & \textbf{1.123 } & \textbf{0.867 } & 0.339  & \textbf{0.318 } & \textbf{0.338 } & 0.323  \\
          & \cellcolor{gray!15}Avg   & \cellcolor{gray!15}0.995  & \cellcolor{gray!15}0.570  & \cellcolor{gray!15}\textbf{0.681 } & \cellcolor{gray!15}\textbf{0.551 } & \cellcolor{gray!15}0.991  & \cellcolor{gray!15}\textbf{0.552 } & \cellcolor{gray!15}\textbf{0.730 } & \cellcolor{gray!15}0.581  & \cellcolor{gray!15}0.898  & \cellcolor{gray!15}0.743  & \cellcolor{gray!15}\textbf{0.887 } & \cellcolor{gray!15}\textbf{0.742 } & \cellcolor{gray!15}1.187  & \cellcolor{gray!15}0.822  & \cellcolor{gray!15}\textbf{0.880 } & \cellcolor{gray!15}\textbf{0.691 } & \cellcolor{gray!15}0.502  & \cellcolor{gray!15}\textbf{0.389 } & \cellcolor{gray!15}\textbf{0.480 } & \cellcolor{gray!15}\textbf{0.389 } \\
    \bottomrule
    \end{tabular}%
    }
  \label{tab:zero_result}%
\end{table*}%

\section{Experiments}

This section comprehensively evaluates the performance of our proposed MEMTS framework on time series forecasting tasks. We first introduce the experimental setup and main results, then analyze the key components of the model, and finally demonstrate its practical effectiveness through case studies.

\subsection{Experimental Setup}

\subsubsection{Datasets.} We evaluate MEMTS on six publicly available time series benchmark datasets: ECL, AustraliaRainfall (denoted as \textbf{AusRain}) \cite{tan2021time}, METR-LA, PEMS04, PEMS08, and Solar. These datasets cover diverse domains including electricity consumption, meteorology, traffic flow, and energy production, exhibiting distinct temporal characteristics, sampling frequencies, and forecasting challenges. For the pretraining dataset, we selected partial sub-datasets of UTSD-12G from the UTSD dataset \cite{DBLP:journals/corr/abs-2402-02368} to pretrain Adaptive Fusion. Adaptive Fusion was not involved in any training of the evaluation datasets.

\subsubsection{Baselines.} We compare MEMTS with five state-of-the-art time series foundation models. Specifically, we utilize the high-efficiency \textit{Chronos-Bolt} family, including its Small and Base variants (denoted as \textbf{Bolt-S} and \textbf{Bolt-B}), alongside \textbf{MOMENT}, \textbf{Moirai}, and \textbf{Sundial}. All baseline methods are evaluated using their official implementations and open-source pretrained weights.

\subsubsection{Implementation.} The evaluation metrics are mean squared error (MSE) and mean absolute error (MAE). To verify the effectiveness of MEMTS across different time horizons, we consider prediction lengths of 96, 192, 336, and 720 steps. Implementation details and the complete experimental setup are provided in Appendix A.1.

\subsection{Zero-shot Augmentation Results}

In this section, we evaluate the effectiveness of MEMTS by integrating it into five frozen state-of-the-art TSFMs: Bolt-S, Bolt-B, Moment, Moirai, and Sundial. The comprehensive comparison results across six datasets and four prediction horizons \\($H \in \{96, 192, 336, 720\}$) are reported in Table \ref{tab:zero_result}.

\begin{figure}[t]
\centering
\includegraphics[width=0.48\textwidth]{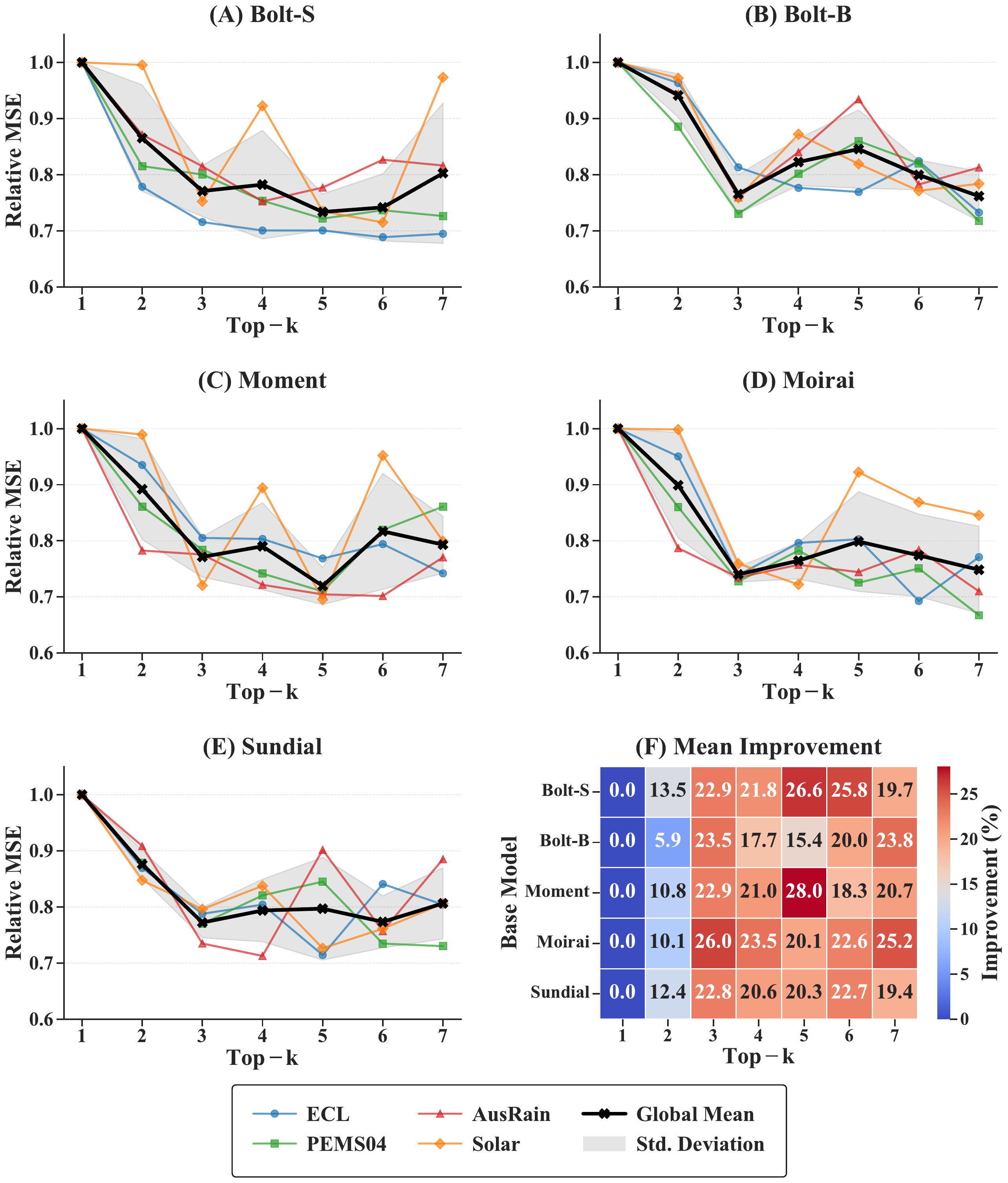}
% \caption{Sensitivity analysis of the retrieval hyperparameter $k$. The results indicate that $k=3$ consistently yields the optimal trade-off between information sufficiency and noise robustness across all datasets.}
\caption{\textbf{Ablation study on retrieval hyperparameter $k$.} 
Subplots \textbf{(A)-(E)} display the Relative MSE trends across four datasets, showing consistent error reduction as $k$ increases. 
Subplot \textbf{(F)} summarizes the mean improvement percentages relative to the $k=1$ baseline. 
The results indicate that performance gains typically saturate around $k \in [3, 5]$, validating that a small set of retrieved prototypes is sufficient for robust adaptation.}
\label{fig:topk-abalation}
\end{figure}

\textbf{Consistent Accuracy Improvement.}
As demonstrated in Table \ref{tab:zero_result}, direct zero-shot inference often suffers from domain misalignment, leading to suboptimal performance, particularly in data-sets with complex temporal dynamics like PEMS04 and PEMS08. However, the integration of MEMTS consistently enhances the forecasting accuracy across all foundation models and datasets. Taking the average MSE as a primary indicator, MEMTS achieves remarkable error reductions. For instance, on the PEMS04 dataset, MEMTS reduces the average MSE of Bolt-S from 1.139 to 0.702, representing a substantial improvement of approximately 38.4\%. Similarly, for Moirai on the ECL dataset, the average MSE drops from 1.020 to 0.671, an improvement of roughly 34.2\%. Even for models that already perform well, such as Sundial on the Solar dataset, MEMTS still provides marginal yet consistent gains (MSE from 0.502 to 0.480).

\textbf{Robustness Across Horizons.}
Beyond average performance, MEMTS exhibits robust effectiveness across varying prediction leng-ths. Whether for short-term forecasting ($H=96$) or long-term forecasting ($H=720$), MEMTS consistently outperforms the baseline.
Notably, in the challenging long-term setting on METR-LA, MEMTS helps Bolt-B reduce the MSE from 2.308 to 2.072. This indicates that the domain-specific prototypes learned by MEMTS are not merely capturing short-term noise but are effectively encoding persistent structural motifs that benefit long-range predictions.

\textbf{Universal Applicability.}
The results further validate the archite-cture-agnostic nature of MEMTS. Regardless of the underlying architecture of the base TSFM—whether it is based on language models (Chronos-Bolt), transformers (Moment, Moirai), or other designs—MEMTS successfully injects domain knowledge to refine the predictions. This confirms that MEMTS serves as a universal plugin for enhancing the adaptability of foundation models without requiring expensive model-specific tuning.

\subsection{Fine-tuning Augmentation Results}
\label{sec:ft_results}

While zero-shot adaptation demonstrates the plug-and-play capability of MEMTS, a critical question remains whether our module can further enhance foundation models that have already been adapted to domain-specific data via fine-tuning. Fine-tuning is a standard approach to mitigate distribution shifts by training the model on the target dataset. To investigate this, we compare the performance of standard Fine-Tuning (\textbf{FT}) against the proposed approach where MEMTS is integrated into the fine-tuned backbone (\textbf{+MEMTS}). Table \ref{tab:ft_result} summarizes the average Mean Squared Error (MSE) and Mean Absolute Error (MAE) across four prediction horizons. \textit{Due to space constraints, detailed results for each horizon are provided in Table \ref{tab:fine_tuned_memts} of Appendix \ref{app:ft_detailed}.}

\begin{table*}[htbp]
  \centering
  \caption{Comparative analysis of forecasting performance between standard Fine-Tuned (FT) foundation models and their MEMTS-augmented counterparts across six datasets. \textbf{Bold} values indicate the best results. Detailed horizon-wise results are available in Appendix \ref{app:ft_detailed}.}
  \resizebox{\textwidth}{!}{
    \begin{tabular}{c|cccc|cccc|cccc|cccc|cccc}
    \toprule
    \multirow{2}[4]{*}{Models} & \multicolumn{4}{c|}{\textbf{Bolt-S}} & \multicolumn{4}{c|}{\textbf{Bolt-B}} & \multicolumn{4}{c|}{\textbf{Moment}} & \multicolumn{4}{c|}{\textbf{Moirai}} & \multicolumn{4}{c}{\textbf{Sundial}} \\
\cmidrule{2-21}          & \multicolumn{2}{c}{FT} & \multicolumn{2}{c|}{+MEMTS} & \multicolumn{2}{c}{FT} & \multicolumn{2}{c|}{+MEMTS} & \multicolumn{2}{c}{FT} & \multicolumn{2}{c|}{+MEMTS} & \multicolumn{2}{c}{FT} & \multicolumn{2}{c|}{+MEMTS} & \multicolumn{2}{c}{FT} & \multicolumn{2}{c}{+MEMTS} \\
    \midrule
    Metric & MSE   & MAE   & MSE   & MAE   & MSE   & MAE   & MSE   & MAE   & MSE   & MAE   & MSE   & MAE   & MSE   & MAE   & MSE   & MAE   & MSE   & MAE   & MSE   & MAE \\
    \midrule
    ECL   & 0.276  & \textbf{0.353 } & \textbf{0.260 } & 0.363  & 0.265  & \textbf{0.345 } & \textbf{0.246 } & 0.349  & 0.885  & 0.774  & \textbf{0.850 } & \textbf{0.757 } & 1.020  & 0.730  & \textbf{0.679 } & \textbf{0.609 } & \textbf{0.195 } & \textbf{0.283 } & \textbf{0.195 } & 0.286  \\
    \midrule
    AusRain & 1.652  & 1.029  & \textbf{1.203 } & \textbf{0.880 } & 1.698  & 1.044  & \textbf{1.245 } & \textbf{0.892 } & 1.101  & 0.858  & \textbf{1.091 } & \textbf{0.855 } & 1.773  & 1.044  & \textbf{1.468 } & \textbf{0.968 } & 1.227  & 0.887  & \textbf{1.213 } & \textbf{0.882 } \\
    \midrule
    METR-LA & 2.092  & 0.966  & \textbf{1.659 } & \textbf{0.851 } & 2.170  & 0.979  & \textbf{1.734 } & \textbf{0.850 } & 1.430  & 0.826  & \textbf{1.393 } & \textbf{0.823 } & 1.716  & 0.810  & \textbf{1.465 } & \textbf{0.770 } & 1.436  & 0.706  & \textbf{1.417 } & \textbf{0.704 } \\
    \midrule
    PEMS04 & 1.139  & 0.777  & \textbf{0.753 } & \textbf{0.649 } & 1.227  & 0.809  & \textbf{0.834 } & \textbf{0.686 } & 1.334  & 0.985  & \textbf{1.236 } & \textbf{0.950 } & 1.456  & 0.898  & \textbf{1.078 } & \textbf{0.795 } & 0.527  & 0.453  & \textbf{0.507 } & \textbf{0.450 } \\
    PEMS08 & 1.210  & 0.815  & \textbf{0.827 } & \textbf{0.690 } & 1.283  & 0.849  & \textbf{0.887 } & \textbf{0.717 } & 1.344  & 0.980  & \textbf{1.246 } & \textbf{0.945 } & 1.386  & 0.859  & \textbf{1.055 } & \textbf{0.774 } & 0.486  & 0.430  & \textbf{0.468 } & \textbf{0.427 } \\
    \midrule
    Solar & 0.995  & \textbf{0.570 } & \textbf{0.781 } & 0.586  & 0.991  & \textbf{0.552 } & \textbf{0.754 } & 0.577  & 0.898  & 0.743  & \textbf{0.850 } & \textbf{0.731 } & 1.187  & 0.822  & \textbf{0.978 } & \textbf{0.763 } & 0.502  & \textbf{0.389 } & \textbf{0.488 } & 0.392  \\
    \bottomrule
    \end{tabular}%
    }
  \label{tab:ft_result}%
\end{table*}%

\textbf{Surpassing the Fine-tuning Baseline.}
The results presented in Table \ref{tab:ft_result} reveal that MEMTS consistently pushes the performance boundary beyond what standard fine-tuning achieves. In the vast majority of cases, the integration of MEMTS yields lower error rates compared to the fine-tuned baselines. This improvement is particularly pronounced in complex datasets characterized by high volatility. For instance, on the PEMS04 traffic dataset, adding MEMTS to the fine-tuned \textbf{Bolt-S} model reduces the MSE from 1.139 to \textbf{0.753}. This corresponds to a remarkable error reduction of approximately \textbf{33.9\%}, suggesting that MEMTS captures localized structural motifs that standard gradient-based adaptation fails to encode effectively. Similarly, for Moirai on the ECL dataset, MEMTS lowers the MSE from 1.020 to \textbf{0.679}, demonstrating its ability to refine predictions even after the model has been tuned on domain knowledge.

\textbf{Complementary Knowledge Representation.}
The experimental evidence suggests that MEMTS and fine-tuning operate via complementary mechanisms. Fine-tuning adjusts the model parameters to align with the general data distribution, whereas MEMTS establishes a dedicated parametric memory to store and retrieve specific prototype patterns. This dual-pathway approach allows the model to leverage both the generalized representations learned via fine-tuning and the explicit historical references provided by MEMTS. Consequently, MEMTS serves as an effective residual correction module, enabling fine-tuned models to achieve superior accuracy without requiring complex architectural modifications.

\subsection{Inference Efficiency Analysis}
\label{sec:efficiency}

Standard RAG approaches rely on explicit nearest neighbor search, creating a bottleneck where latency grows linearly with the knowledge base size ($N_{KB}$). In contrast, MEMTS employs the \textit{Knowledge Persistence Module (KPM)} to internalize patterns into fixed-size parameters, effectively decoupling inference speed from storage volume. 
To validate this, we benchmarked KPM against a standard IVFFlat RAG baseline on an NVIDIA H200 GPU (batch size 32). As shown in Table \ref{tab:latency}, KPM maintains a stable, ultra-low latency of $\approx$ \textbf{1.1 ms} across all configurations, independent of the datastore scale. Conversely, RAG latency degrades significantly, averaging $\sim$160 ms for a 100K entry base. Consequently, KPM achieves an industrial-grade throughput ($>$29,000 samples/s) and a massive \textbf{712.2$\times$} speedup at the 500K scale. Furthermore, unlike the fluctuating search times of RAG, KPM ensures deterministic execution via optimized matrix multiplications. These results confirm MEMTS as a robust solution for latency-sensitive applications like high-frequency trading, eliminating the retrieval overhead inherent in traditional RAG.

\begin{table}[htbp]
\centering
\caption{Inference efficiency comparison between our proposed Knowledge Persistence Module (KPM) and standard RAG. The evaluation measures latency (ms) and throughput (samples/s) with a batch size of 32. Speedup factors are reported across varying Knowledge Base (KB) sizes ($10\text{K}, 100\text{K}, 500\text{K}$).}
\label{tab:latency}
\resizebox{0.48\textwidth}{!}{
\begin{tabular}{@{}lcccccccc@{}}
\toprule
\multirow{2}{*}{\textbf{Config}} & \multicolumn{2}{c}{\textbf{Latency (ms)}} & \multicolumn{3}{c}{\textbf{Speedup ($\times$) vs. RAG}} & \multicolumn{2}{c}{\textbf{Throughput}} \\
\cmidrule(lr){2-3} \cmidrule(lr){4-6} \cmidrule(lr){7-8}
($L_{in}$-$L_{out}$) & \textbf{KPM} & RAG\scriptsize{@100K} & @10K & @100K & @500K & \textbf{KPM} & RAG \\
\midrule
96-96   & \textbf{1.10}$\pm$0.08 & 167.24 & 34.5 & 152.1 & \textbf{712.2} & \textbf{29,096} & 191 \\
336-192 & \textbf{1.10}$\pm$0.07 & 168.86 & 33.0 & 153.9 & 485.2 & \textbf{29,159} & 190 \\
336-336 & \textbf{1.12}$\pm$0.07 & 158.52 & 32.7 & 142.2 & 597.2 & \textbf{28,695} & 202 \\
512-720 & \textbf{1.16}$\pm$0.06 & 142.12 & 29.9 & 122.7 & 448.3 & \textbf{27,619} & 225 \\
\bottomrule
\end{tabular}
}
\end{table}

\begin{figure}[t]
\centering
\includegraphics[width=0.48\textwidth]{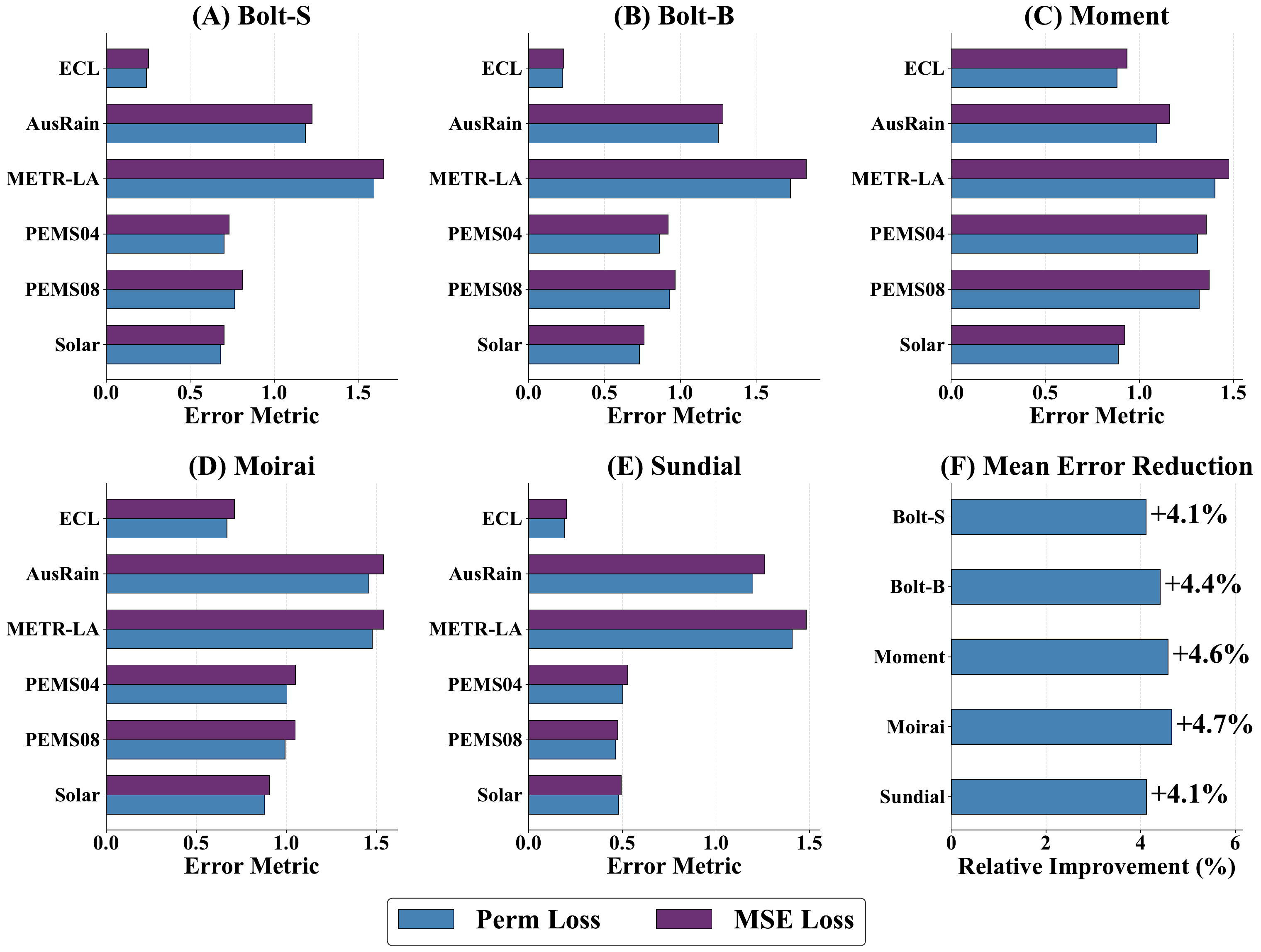}
\caption{Impact of Perm Loss on foundation model performance across diverse time-series datasets. Subplots (A) through (E) show that integrating Perm Loss consistently yields lower errors compared to the MSE baseline in all tested scenarios. Subplot (F) quantifies the average improvement.}
\label{fig:loss-abalation}
\end{figure}

\begin{figure}[t]
\centering
\includegraphics[width=0.48\textwidth]{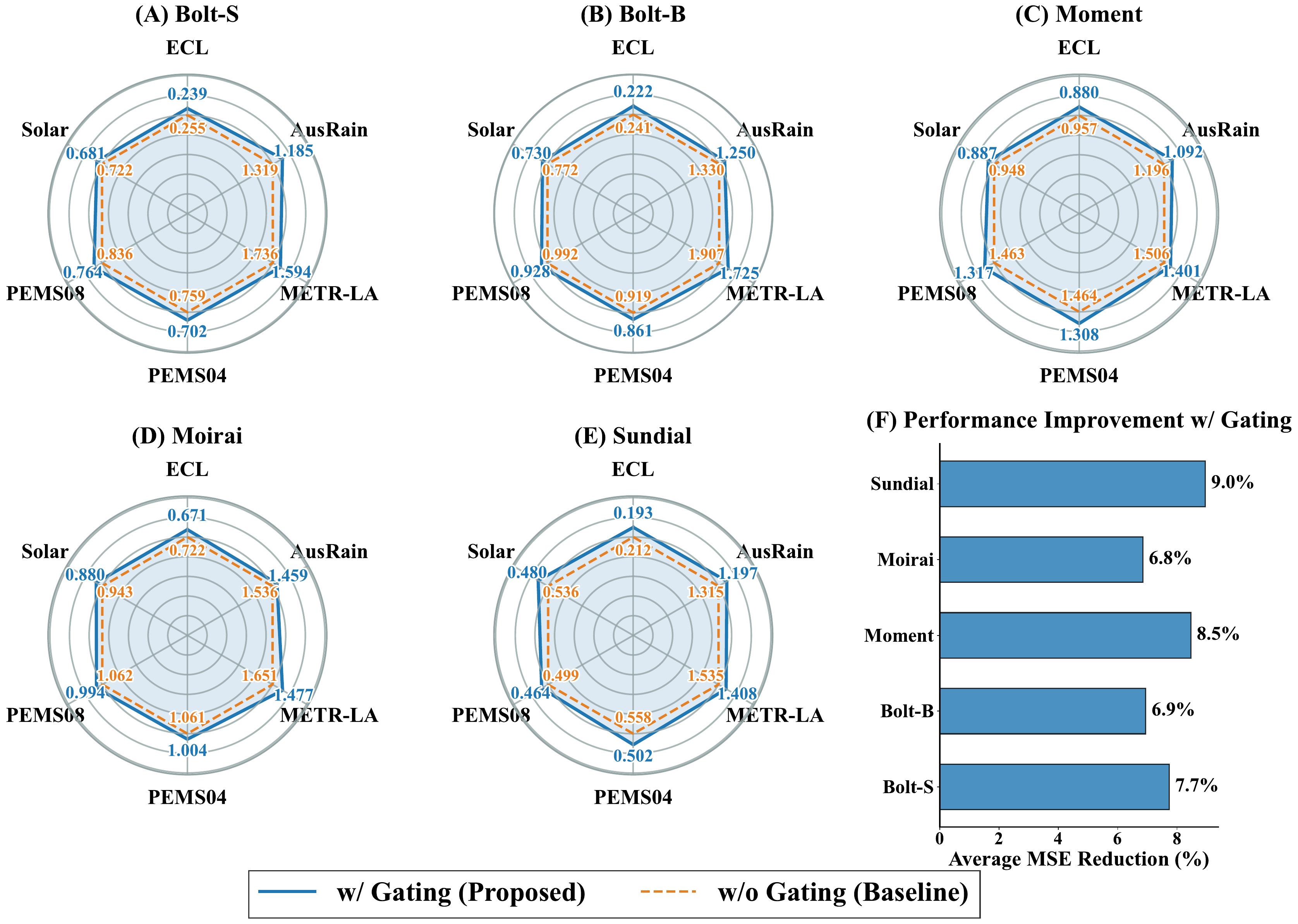}
\caption{Impact of the Gating Mechanism. (A)-(E) Radar charts showing normalized MSE improvements. Blue solid lines (w/ Gating) enveloping orange dashed lines (w/o Gating) indicate superior performance. Vertices display absolute MSE values. (F) Average percentage of MSE reduction achieved by gating across all datasets.}
\label{fig:fusion-gate-abalation}
\end{figure}

\subsection{Ablation Study}

To validate the effectiveness of the key components in MEMTS, we conducted a comprehensive ablation study focusing on two aspects: the internal design choices of the MEMTS module (retrieval size $k$ and loss function) and the integration strategy (Adaptive Fusion).

% \subsubsection{Impact of Retrieval Hyperparameter $k$}
% The number of retrieved prototypes, $k$, determines the richness of the historical context fused into the forecast. We evaluated the performance of MEMTS with $k$ ranging from 1 to 10 across all five foundation models. As illustrated in Figure \ref{fig:topk-abalation}, the performance initially improves as $k$ increases, benefiting from richer domain knowledge. However, beyond a certain threshold (typically $k=3$), the error rates plateau or even slightly increase, likely due to the introduction of irrelevant or noisy prototypes. Consequently, we observe that $k=3$ serves as a robust ``sweet spot'' across diverse datasets and architectures, providing sufficient context while maintaining retrieval precision.

\subsubsection{Impact of Retrieval Hyperparameter $k$}
The hyperparameter $k$ governs the capacity of the memory retrieval, determining how many historical prototypes are fused to guide the forecast. To investigate its sensitivity, we evaluated the performance of MEMTS with $k$ ranging from 1 to 7 across five foundation models (Bolt-S, Bolt-B, Moment, Moirai, and Sundial) on four representative datasets. 

As visualized in Figure \ref{fig:topk-abalation} (A)-(E), the \textbf{Relative MSE} curves exhibit a consistent downward trend as $k$ increases from 1, demonstrating that a single reference prototype is often insufficient to capture complex temporal dynamics. Notably, the most significant performance gains occur typically between $k=1$ and $k=3$. As shown in the heatmap (Figure \ref{fig:topk-abalation} F), MEMTS achieves substantial improvements over the baseline across all architectures. However, beyond $k=5$, the marginal gains diminish, and in some cases (e.g., Sundial on Solar), the error rates plateau or slightly fluctuate due to the potential inclusion of less relevant patterns. Consequently, we identify $k \in [3, 5]$ as the optimal range, balancing retrieval precision with computational efficiency, and select $k=3$ as the default setting for our main experiments.

\subsubsection{Effectiveness of Permutation Loss}

A core contribution of MEMTS is the training of the Knowledge Persistence Module (KPM) using a specialized Permutation Loss ($\mathcal{L}_{perm}$) rather than a standard reconstruction loss (e.g., MSE). To verify its necessity, we trained variants of MEMTS using standard MSE loss while keeping other architectures unchanged. 

Figure \ref{fig:loss-abalation} presents the comparison across five foundation models. The results consistently show that models trained with $\mathcal{L}_{perm}$ (blue bars) achieve lower forecasting errors compared to their MSE counterparts (purple bars). As summarized in Figure \ref{fig:loss-abalation} (F), the permutation loss yields an average relative improvement ranging from \textbf{4.1\%} to \textbf{4.7\%} across different backbones. For instance, on the challenging PEMS04 dataset, \textbf{Bolt-S} trained with Perm Loss significantly outperforms the MSE variant. This suggests that explicitly optimizing the retrieval rank order via $\mathcal{L}_{perm}$ encourages the learned prototypes to capture more discriminative structural motifs, whereas pointwise reconstruction loss often leads to overly smoothed representations.

\subsubsection{Necessity of the Gating Mechanism}
Finally, we investigate the role of the Adaptive Fusion module, specifically the gating mechanism that dynamically weights the contribution of the retrieved memory. We compared our proposed gated fusion against a baseline where the memory output is directly added to the TSFM prediction without adaptive weighting (w/o Gating). The radar charts in Figure \ref{fig:fusion-gate-abalation} clearly demonstrate that removing the gate leads to a universal performance degradation (represented by the orange dashed lines being consistently enclosed by the blue solid lines). The quantitative analysis in Figure \ref{fig:fusion-gate-abalation}(F) reveals that the gating mechanism is responsible for an average error reduction of approximately 7.4\% to 9.8\%. This confirms that blindly injecting retrieved knowledge can be detrimental, and the adaptive gate is crucial for filtering noise and ensuring that the memory module intervenes only when beneficial.

\begin{table}[htbp]
  \centering
  \caption{Performance comparison with representative adaptation methods (\textbf{RAG} and \textbf{DAPT}) under the condition of $H=96$. \textbf{Bold} indicates the best performance, and \underline{underlined} indicates the second best. MEMTS achieves competitive accuracy while maintaining superior efficiency.}
  \resizebox{0.48\textwidth}{!}{
    \begin{tabular}{c|cc|cc|cc|cc|cc|cc}
    \toprule
    \multirow{2}[4]{*}{Metric} & \multicolumn{2}{c|}{ECL} & \multicolumn{2}{c|}{AusRain} & \multicolumn{2}{c|}{METR-LA} & \multicolumn{2}{c|}{PEMS04} & \multicolumn{2}{c|}{PEMS08} & \multicolumn{2}{c}{Solar} \\
\cmidrule{2-13}          & MSE   & MAE   & MSE   & MAE   & MSE   & MAE   & MSE   & MAE   & MSE   & MAE   & MSE   & MAE \\
    \midrule
    Bolt-S & 0.360  & \underline{0.411}  & 2.225  & 1.198  & 2.406  & 1.092  & 2.214  & 1.188  & 2.204  & 1.189  & 1.257  & 0.699  \\
    +RAG  & 0.357 & \textbf{0.408} & 2.193 & 1.189 & 2.354 & 1.083 & 2.177 & 1.178 & 2.165 & 1.178 & 1.091 & 0.641 \\
    +DAPT & \underline{0.333} & 0.420  & \underline{1.340}  & \underline{0.921} & \underline{1.460}  & \underline{0.788} & \textbf{0.767} & \textbf{0.664} & \textbf{0.821} & \textbf{0.708} & \textbf{0.467} & \underline{0.470} \\
    +MEMTS & \textbf{0.327}  & 0.418  & \textbf{1.337}  & \textbf{0.914}  & \textbf{1.440}  & \textbf{0.779}  & \underline{0.775}  & \underline{0.677}  & \underline{0.827}  & \underline{0.715}  & \underline{0.472}  & \textbf{0.462}  \\
    \midrule
    Bolt-B & 0.329  & 0.389  & 2.077  & 1.150  & 2.436  & 1.089  & 2.189  & 1.177  & 2.186  & 1.185  & 1.277  & 0.685  \\
    +RAG  & 0.326 & 0.386 & 2.048 & 1.141 & 2.384 & 1.080 & 2.153 & 1.167 & 2.147 & 1.175 & 1.118 & 0.629 \\
    +DAPT & \underline{0.270}  & \textbf{0.351} & \underline{1.329} & \textbf{0.901} & \underline{1.779} & \textbf{0.872} & \textbf{1.145} & \textbf{0.828} & \underline{1.194} & \textbf{0.859} & \underline{0.717} & \underline{0.556} \\
    +MEMTS & \textbf{0.266}  & \underline{0.354}  & \textbf{1.321}  & \underline{0.905}  & \textbf{1.750}  & \underline{0.873}  & \underline{1.156}  & \underline{0.836}  & \textbf{1.191}  & \underline{0.867}  & \textbf{0.715}  & \textbf{0.547}  \\
    \midrule
    Moment & 0.867  & 0.767  & 1.170  & 0.880  & 1.386  & 0.765  & 2.045  & 1.238  & 2.075  & 1.243  & 1.071  & 0.782  \\
    +RAG  & 0.865 & \underline{0.766} & \underline{1.167} & 0.879 & 1.358 & 0.762 & 2.014 & 1.229 & 2.043 & 1.233 & \textbf{0.915} & \textbf{0.734} \\
    +DAPT & \textbf{0.857} & \textbf{0.756} & 1.176 & \underline{0.874} & \textbf{1.324} & \textbf{0.748} & \textbf{1.930}  & \underline{1.219} & \underline{2.012} & \textbf{1.214} & \underline{1.026} & \underline{0.767} \\
    +MEMTS & \underline{0.860}  & \textbf{0.756}  & \textbf{1.153}  & \textbf{0.874}  & \underline{1.346}  & \underline{0.762}  & \underline{1.961}  & \textbf{1.213}  & \textbf{1.989}  & \underline{1.218}  & 1.044  & 0.771  \\
    \midrule
    Moirai & 1.412  & 0.906  & 2.261  & 1.176  & 1.570  & 0.758  & 2.092  & 1.188  & 1.881  & 1.094  & 1.440  & 0.929  \\
    +RAG  & 1.405 & 0.904 & 2.254 & 1.174 & 1.540 & 0.755 & 2.061 & 1.179 & 1.852 & 1.086 & 1.256 & 0.877 \\
    +DAPT & \underline{0.657} & \underline{0.625} & \underline{1.534} & \underline{0.995} & \textbf{1.193} & \underline{0.672} & \underline{0.828} & \underline{0.732} & \textbf{0.829} & \underline{0.739} & \textbf{0.544} & \textbf{0.557} \\
    +MEMTS & \textbf{0.644}  & \textbf{0.624}  & \textbf{1.506}  & \textbf{0.981}  & \underline{1.217}  & \textbf{0.666}  & \textbf{0.822}  & \textbf{0.731}  & \underline{0.837}  & \textbf{0.735}  & \underline{0.553}  & \underline{0.558}  \\
    \midrule
    Sundial & 0.207  & \textbf{0.286}  & 1.240  & 0.878  & 1.368  & \underline{0.684}  & 1.481  & 0.943  & 1.403  & 0.910  & 0.907  & 0.594  \\
    +RAG  & 0.207 & \underline{0.287} & 1.227 & 0.874 & 1.338 & 0.682 & 1.456 & 0.936 & 1.376 & 0.901 & \textbf{0.771} & \textbf{0.552} \\
    +DAPT & \textbf{0.204} & 0.292 & \underline{1.217} & \textbf{0.857} & \textbf{1.327} & \textbf{0.665} & \underline{1.408} & \underline{0.923} & \textbf{1.303} & \underline{0.884} & 0.842 & \underline{0.566} \\
    +MEMTS & \textbf{0.204}  & \underline{0.287}  & \textbf{1.199}  & \underline{0.865}  & \textbf{1.327}  & \textbf{0.665}  & \textbf{1.385}  & \textbf{0.910}  & \underline{1.314}  & \textbf{0.879}  & \underline{0.838}  & 0.574  \\
    \bottomrule
    \end{tabular}%
    }
  \label{tab:raf-dapt-memts}%
\end{table}%

\subsection{Comparison with RAG and DAPT}
To position MEMTS within the broader landscape of domain adaptation, we compare it against two representative approaches: \textbf{Retrieval-Augmented Generation (RAG)} and \textbf{Domain-Adaptive Pretraining (DAPT)}. Specifically, for the RAG paradigm, we compare with \textbf{RAF} (Retrieval Augmented Forecasting) \cite{tire2024retrieval}, a recent retrieval-augmented method for zero-shot time series forecasting. DAPT represents the full-parameter adaptation approach, involving continued pretraining of the foundation model on domain-specific data. Table \ref{tab:raf-dapt-memts} details the performance comparison on six datasets.

\textbf{Competitive Accuracy.}
The results in Table \ref{tab:raf-dapt-memts} demonstrate that MEMTS achieves forecasting accuracy that is highly competitive with both RAG and DAPT. 
For instance, on the \textbf{Bolt-B} backbone, MEMTS achieves an MSE of \textbf{0.266} on the ECL dataset, notably outperforming both the fine-tuning-based DAPT (0.270) and the retrieval-based RAG (0.326). 
Overall, MEMTS exhibits robust adaptability across diverse domains, suggesting that its internal parametric memory successfully captures domain-specific nuances—achieving the precision of retrieval and fine-tuning methods.

\textbf{Superior Efficiency and Deployment Feasibility.}
While MEMTS delivers accuracy on par with these baselines, its primary advantage lies in its architectural efficiency:
\begin{itemize}[leftmargin=*]
    \item \textbf{Vs. RAG:} RAG incurs a significant inference bottleneck due to the necessity of searching through massive external databases for every query. In contrast, MEMTS offers a high-efficiency solution where retrieval is replaced by a lightweight forward pass, eliminating the need for maintaining non-parametric datastores.
    \item \textbf{Vs. DAPT:} DAPT typically requires maintaining separate, fully fine-tuned copies of the foundation model for each domain, leading to high storage costs. MEMTS, however, employs a compact memory module, enabling more efficient multi-domain serving compared to full-model adaptation.
\end{itemize}
In summary, MEMTS represents a Pareto-optimal solution, bridging the gap between high-precision adaptation and practical deployment requirements.

\section{Conclusion}
We propose MEMTS, a plug-and-play parameterized memory framework that addresses the dilemma between domain adaptation and catastrophic forgetting in time series foundation models. By encoding rare yet critical temporal patterns into lightweight memory modules, MEMTS enables efficient knowledge injection without modifying backbone parameters or incurring online retrieval overhead. The proposed KPM models multimodal future evolution through unordered multi-future prediction, while an adaptive gated fusion mechanism ensures stable behavior in stationary regions and accurate responses to abrupt changes. Extensive experiments show that MEMTS consistently improves multiple mainstream TSFMs in both zero-shot and fine-tuning settings, significantly enhancing forecasting accuracy while maintaining inference efficiency comparable to the base models. Overall, MEMTS provides an efficient, generalizable, and interpretable paradigm for enhancing time series foundation models in dynamic real-world environments.

\clearpage

% \begin{acks}
% \end{acks}

\bibliographystyle{ACM-Reference-Format}
\bibliography{cite}

\clearpage
\appendix

\section{Dataset Details}
\label{app:datasets}

We evaluate MEMTS on six widely used real-world benchmarks covering diverse domains such as energy, meteorology, and traffic. All datasets were standardized using z-score normalization. We followed the standard chronological split of 70\%/10\%/20\% for training, validation, and testing, respectively. The detailed statistics are summarized in Table \ref{tab:dataset_stats}.

\begin{itemize}
    \item \textbf{ECL (Electricity Consuming Load):} Records the hourly electricity consumption of 321 clients from 2012 to 2014. It captures long-term cyclic patterns in power usage.
    \item \textbf{AusRain:} A meteorological dataset containing daily rainfall records from various weather stations across Australia, characterized by high volatility and sparse non-zero values.
    \item \textbf{METR-LA \& PEMS04/08:} Large-scale traffic forecasting datasets collected from highway sensors in Los Angeles and the San Francisco Bay Area. These datasets exhibit strong daily and weekly periodicities but are prone to distribution shifts due to traffic events.
    \item \textbf{Solar:} Records the solar power production from 137 PV plants in Alabama State, sampled every 10 minutes. This dataset is highly dependent on daylight cycles and weather conditions.
\end{itemize}

\begin{table}[h]
    \centering
    \caption{Detailed Statistics of the Evaluated Datasets.}
    \label{tab:dataset_stats}
    \resizebox{0.48\textwidth}{!}{
    \begin{tabular}{lcccc}
    \toprule
    \textbf{Dataset} & \textbf{Frequency} & \textbf{Variates} & \textbf{Timesteps} & \textbf{Domain} \\
    \midrule
    ECL & 1 Hour & 321 & 26,304 & Electricity \\
    AusRain & Hourly & 3 & 24 & Environment \\
    METR-LA & 5 Min & 207 & 34,272 & Traffic \\
    PEMS04 & 5 Min & 307 & 16,992 & Traffic \\
    PEMS08 & 5 Min & 170 & 17,856 & Traffic \\
    Solar & 10 Min & 137 & 52,560 & Energy \\
    \bottomrule
    \end{tabular}}
\end{table}

\section{Baseline Implementations}
\label{app:baselines}

To ensure a rigorous evaluation, we benchmark MEMTS against five state-of-the-art Time Series Foundation Models (TSFMs). All baselines were evaluated using their official open-source implementations and pre-trained checkpoints.

\textbf{Chronos-Bolt (Small \& Base):} A family of efficient pretrained probabilistic forecasting models based on the T5 encoder-decoder architecture. Unlike the original Chronos which tokenizes values via quantization, Chronos-Bolt utilizes a patch-based approach for faster inference. We use the chronos-bolt-small and chronos-bolt-base variants.

\textbf{MOMENT:} An open-source family of foundation models pre-trained on the Monash Time Series Repository. We employ the \texttt{MOMENT-1-large} checkpoint, which utilizes a T5-based backbone and is designed for low-resource adaptation.

\textbf{Moirai:} A Universal Time Series Forecasting approach capable of handling diverse frequencies. We use the \texttt{moirai-1.0-R-base} variant, which employs a patch size of 64 and varying attention mechanisms to capture multi-scale dependencies.

\textbf{Sundial:} A recent foundation model designed for continuous-time modeling. We utilize its default configuration as provided in the official repository, ensuring consistent preprocessing and evaluation protocols.

\begin{table*}[htbp]
  \centering
  \caption{MEMTS enhances fine-tuned time series foundation models: Inserting MEMTS into fine-tuned TSFMs consistently reduces forecasting error across 6 datasets and 4 horizons. Gray rows denote average performance.}
  \resizebox{\textwidth}{!}{
    \begin{tabular}{c|c|cccc|cccc|cccc|cccc|cccc}
    \toprule
    \multicolumn{2}{c|}{\multirow{2}[4]{*}{Models}} & \multicolumn{4}{c}{\textbf{Bolt-S}}    & \multicolumn{4}{c}{\textbf{Bolt-B}}    & \multicolumn{4}{c}{\textbf{Moment}}    & \multicolumn{4}{c|}{\textbf{Moirai}}   & \multicolumn{4}{c}{\textbf{Sundial}} \\
\cmidrule{3-22}    \multicolumn{2}{c|}{} & \multicolumn{2}{c}{FT} & \multicolumn{2}{c|}{+MEMTS} & \multicolumn{2}{c}{FT} & \multicolumn{2}{c|}{+MEMTS} & \multicolumn{2}{c}{FT} & \multicolumn{2}{c|}{+MEMTS} & \multicolumn{2}{c}{FT} & \multicolumn{2}{c|}{+MEMTS} & \multicolumn{2}{c}{FT} & \multicolumn{2}{c|}{+MEMTS} \\
    \midrule
    \multicolumn{2}{c|}{Metric} & MSE   & MAE   & MSE   & MAE   & MSE   & MAE   & MSE   & MAE   & MSE   & MAE   & MSE   & MAE   & MSE   & MAE   & MSE   & MAE   & MSE   & MAE   & MSE   & MAE \\
    \midrule
    \multirow{5}[2]{*}{ECL} & 96    & 0.360  & \textbf{0.411 } & \textbf{0.314 } & 0.412  & 0.329  & \textbf{0.389 } & \textbf{0.311 } & 0.409  & 0.867  & 0.767  & \textbf{0.807 } & \textbf{0.739 } & 1.412  & 0.906  & \textbf{0.665 } & \textbf{0.628 } & 0.207  & \textbf{0.286 } & \textbf{0.206 } & 0.289  \\
          & 192   & 0.242  & 0.333  & \textbf{0.213 } & \textbf{0.324 } & 0.241  & 0.333  & \textbf{0.201 } & \textbf{0.311 } & 0.877  & 0.772  & \textbf{0.851 } & \textbf{0.761 } & \textbf{0.191 } & \textbf{0.277 } & 0.207  & 0.305  & \textbf{0.167 } & \textbf{0.260 } & 0.168  & 0.264  \\
          & 336   & \textbf{0.238 } & \textbf{0.327 } & 0.251  & 0.358  & 0.236  & \textbf{0.325 } & \textbf{0.227 } & 0.341  & 0.891  & 0.777  & \textbf{0.854 } & \textbf{0.757 } & 1.244  & 0.867  & \textbf{0.851 } & \textbf{0.723 } & \textbf{0.185 } & \textbf{0.278 } & 0.186  & 0.280  \\
          & 720   & 0.264  & \textbf{0.340 } & \textbf{0.263 } & 0.356  & 0.253  & \textbf{0.331 } & \textbf{0.243 } & 0.336  & 0.905  & 0.779  & \textbf{0.886 } & \textbf{0.769 } & 1.233  & 0.869  & \textbf{0.994 } & \textbf{0.779 } & \textbf{0.220 } & \textbf{0.308 } & \textbf{0.220 } & 0.310  \\
          & \cellcolor[rgb]{ .925,  .925,  .925}Avg & \cellcolor[rgb]{ .925,  .925,  .925}0.276  & \cellcolor[rgb]{ .925,  .925,  .925}\textbf{0.353 } & \cellcolor[rgb]{ .925,  .925,  .925}\textbf{0.260 } & \cellcolor[rgb]{ .925,  .925,  .925}0.363  & \cellcolor[rgb]{ .925,  .925,  .925}0.265  & \cellcolor[rgb]{ .925,  .925,  .925}\textbf{0.345 } & \cellcolor[rgb]{ .925,  .925,  .925}\textbf{0.246 } & \cellcolor[rgb]{ .925,  .925,  .925}0.349  & \cellcolor[rgb]{ .925,  .925,  .925}0.885  & \cellcolor[rgb]{ .925,  .925,  .925}0.774  & \cellcolor[rgb]{ .925,  .925,  .925}\textbf{0.850 } & \cellcolor[rgb]{ .925,  .925,  .925}\textbf{0.757 } & \cellcolor[rgb]{ .925,  .925,  .925}1.020  & \cellcolor[rgb]{ .925,  .925,  .925}0.730  & \cellcolor[rgb]{ .925,  .925,  .925}\textbf{0.679 } & \cellcolor[rgb]{ .925,  .925,  .925}\textbf{0.609 } & \cellcolor[rgb]{ .925,  .925,  .925}\textbf{0.195 } & \cellcolor[rgb]{ .925,  .925,  .925}\textbf{0.283 } & \cellcolor[rgb]{ .925,  .925,  .925}\textbf{0.195 } & \cellcolor[rgb]{ .925,  .925,  .925}0.286  \\
    \midrule
    \multirow{5}[2]{*}{AusRain} & 96    & 2.225  & 1.198  & \textbf{1.241 } & \textbf{0.884 } & 2.077  & 1.150  & \textbf{1.199 } & \textbf{0.865 } & 1.170  & 0.880  & \textbf{1.128 } & \textbf{0.866 } & 2.261  & 1.176  & \textbf{1.503 } & \textbf{0.976 } & 1.240  & 0.878  & \textbf{1.206 } & \textbf{0.868 } \\
          & 192   & 1.653  & 1.036  & \textbf{1.189 } & \textbf{0.874 } & 1.676  & 1.041  & \textbf{1.212 } & \textbf{0.880 } & 1.086  & 0.853  & \textbf{1.085 } & \textbf{0.853 } & 1.123  & 0.849  & \textbf{1.024 } & \textbf{0.817 } & 1.213  & 0.880  & \textbf{1.198 } & \textbf{0.875 } \\
          & 336   & 1.421  & 0.960  & \textbf{1.162 } & \textbf{0.868 } & 1.461  & 0.972  & \textbf{1.158 } & \textbf{0.868 } & \textbf{1.075 } & \textbf{0.849 } & 1.076  & 0.850  & 1.950  & 1.095  & \textbf{1.713 } & \textbf{1.050 } & 1.226  & 0.890  & \textbf{1.218 } & \textbf{0.888 } \\
          & 720   & 1.309  & 0.923  & \textbf{1.219 } & \textbf{0.892 } & 1.578  & 1.014  & \textbf{1.410 } & \textbf{0.956 } & \textbf{1.073 } & \textbf{0.848 } & 1.076  & 0.850  & 1.756  & 1.055  & \textbf{1.633 } & \textbf{1.027 } & 1.230  & \textbf{0.897 } & \textbf{1.227 } & \textbf{0.897 } \\
          & \cellcolor[rgb]{ .925,  .925,  .925}Avg & \cellcolor[rgb]{ .925,  .925,  .925}1.652  & \cellcolor[rgb]{ .925,  .925,  .925}1.029  & \cellcolor[rgb]{ .925,  .925,  .925}\textbf{1.203 } & \cellcolor[rgb]{ .925,  .925,  .925}\textbf{0.880 } & \cellcolor[rgb]{ .925,  .925,  .925}1.698  & \cellcolor[rgb]{ .925,  .925,  .925}1.044  & \cellcolor[rgb]{ .925,  .925,  .925}\textbf{1.245 } & \cellcolor[rgb]{ .925,  .925,  .925}\textbf{0.892 } & \cellcolor[rgb]{ .925,  .925,  .925}1.101  & \cellcolor[rgb]{ .925,  .925,  .925}0.858  & \cellcolor[rgb]{ .925,  .925,  .925}\textbf{1.091 } & \cellcolor[rgb]{ .925,  .925,  .925}\textbf{0.855 } & \cellcolor[rgb]{ .925,  .925,  .925}1.773  & \cellcolor[rgb]{ .925,  .925,  .925}1.044  & \cellcolor[rgb]{ .925,  .925,  .925}\textbf{1.468 } & \cellcolor[rgb]{ .925,  .925,  .925}\textbf{0.968 } & \cellcolor[rgb]{ .925,  .925,  .925}1.227  & \cellcolor[rgb]{ .925,  .925,  .925}0.887  & \cellcolor[rgb]{ .925,  .925,  .925}\textbf{1.213 } & \cellcolor[rgb]{ .925,  .925,  .925}\textbf{0.882 } \\
    \midrule
    \multirow{5}[2]{*}{METR-LA} & 96    & 2.406  & 1.092  & \textbf{1.575 } & \textbf{0.811 } & 2.436  & 1.089  & \textbf{1.625 } & \textbf{0.819 } & 1.386  & 0.765  & \textbf{1.313 } & \textbf{0.746 } & 1.570  & 0.758  & \textbf{1.252 } & \textbf{0.671 } & 1.368  & 0.684  & \textbf{1.338 } & \textbf{0.677 } \\
          & 192   & 1.905  & 0.928  & \textbf{1.492 } & \textbf{0.805 } & 2.018  & 0.958  & \textbf{1.566 } & \textbf{0.809 } & 1.281  & 0.789  & \textbf{1.255 } & \textbf{0.788 } & 1.484  & \textbf{0.645 } & \textbf{1.326 } & 0.680  & 1.336  & 0.654  & \textbf{1.318 } & \textbf{0.652 } \\
          & 336   & 1.836  & 0.873  & \textbf{1.562 } & \textbf{0.813 } & 1.916  & 0.889  & \textbf{1.629 } & \textbf{0.827 } & 1.384  & \textbf{0.820 } & \textbf{1.353 } & 0.833  & 1.754  & 0.865  & \textbf{1.461 } & \textbf{0.802 } & 1.412  & \textbf{0.695 } & \textbf{1.398 } & 0.696  \\
          & 720   & 2.223  & \textbf{0.972 } & \textbf{2.006 } & 0.973  & 2.308  & 0.982  & \textbf{2.114 } & \textbf{0.946 } & 1.671  & 0.930  & \textbf{1.650 } & \textbf{0.926 } & 2.055  & 0.974  & \textbf{1.822 } & \textbf{0.928 } & 1.628  & 0.792  & \textbf{1.615 } & \textbf{0.791 } \\
          & \cellcolor[rgb]{ .925,  .925,  .925}Avg & \cellcolor[rgb]{ .925,  .925,  .925}2.092  & \cellcolor[rgb]{ .925,  .925,  .925}0.966  & \cellcolor[rgb]{ .925,  .925,  .925}\textbf{1.659 } & \cellcolor[rgb]{ .925,  .925,  .925}\textbf{0.851 } & \cellcolor[rgb]{ .925,  .925,  .925}2.170  & \cellcolor[rgb]{ .925,  .925,  .925}0.979  & \cellcolor[rgb]{ .925,  .925,  .925}\textbf{1.734 } & \cellcolor[rgb]{ .925,  .925,  .925}\textbf{0.850 } & \cellcolor[rgb]{ .925,  .925,  .925}1.430  & \cellcolor[rgb]{ .925,  .925,  .925}0.826  & \cellcolor[rgb]{ .925,  .925,  .925}\textbf{1.393 } & \cellcolor[rgb]{ .925,  .925,  .925}\textbf{0.823 } & \cellcolor[rgb]{ .925,  .925,  .925}1.716  & \cellcolor[rgb]{ .925,  .925,  .925}0.810  & \cellcolor[rgb]{ .925,  .925,  .925}\textbf{1.465 } & \cellcolor[rgb]{ .925,  .925,  .925}\textbf{0.770 } & \cellcolor[rgb]{ .925,  .925,  .925}1.436  & \cellcolor[rgb]{ .925,  .925,  .925}0.706  & \cellcolor[rgb]{ .925,  .925,  .925}\textbf{1.417 } & \cellcolor[rgb]{ .925,  .925,  .925}\textbf{0.704 } \\
    \midrule
    \multirow{5}[2]{*}{PEMS04} & 96    & 2.214  & 1.188  & \textbf{1.064 } & \textbf{0.808 } & 2.189  & 1.177  & \textbf{1.056 } & \textbf{0.793 } & 2.045  & 1.238  & \textbf{1.731 } & \textbf{1.131 } & 2.092  & 1.188  & \textbf{1.007 } & \textbf{0.820 } & 1.481  & 0.943  & \textbf{1.400 } & \textbf{0.918 } \\
          & 192   & 0.926  & 0.718  & \textbf{0.681 } & \textbf{0.631 } & 1.127  & 0.794  & \textbf{0.809 } & \textbf{0.695 } & 1.132  & 0.918  & \textbf{1.089 } & \textbf{0.900 } & \textbf{0.260 } & \textbf{0.308 } & 0.277  & 0.367  & \textbf{0.191 } & \textbf{0.273 } & 0.193  & 0.278  \\
          & 336   & 0.840  & 0.675  & \textbf{0.736 } & \textbf{0.649 } & 0.975  & 0.725  & \textbf{0.852 } & \textbf{0.702 } & 1.057  & 0.882  & \textbf{1.032 } & \textbf{0.870 } & 1.737  & 1.043  & \textbf{1.442 } & \textbf{0.970 } & \textbf{0.214 } & \textbf{0.292 } & 0.215  & 0.296  \\
          & 720   & 0.574  & 0.527  & \textbf{0.531 } & \textbf{0.510 } & 0.616  & \textbf{0.538 } & 0.621  & 0.556  & 1.101  & 0.902  & \textbf{1.091 } & \textbf{0.898 } & 1.736  & 1.052  & \textbf{1.586 } & \textbf{1.022 } & \textbf{0.220 } & \textbf{0.305 } & 0.221  & 0.307  \\
          & \cellcolor[rgb]{ .925,  .925,  .925}Avg & \cellcolor[rgb]{ .925,  .925,  .925}1.139  & \cellcolor[rgb]{ .925,  .925,  .925}0.777  & \cellcolor[rgb]{ .925,  .925,  .925}\textbf{0.753 } & \cellcolor[rgb]{ .925,  .925,  .925}\textbf{0.649 } & \cellcolor[rgb]{ .925,  .925,  .925}1.227  & \cellcolor[rgb]{ .925,  .925,  .925}0.809  & \cellcolor[rgb]{ .925,  .925,  .925}\textbf{0.834 } & \cellcolor[rgb]{ .925,  .925,  .925}\textbf{0.686 } & \cellcolor[rgb]{ .925,  .925,  .925}1.334  & \cellcolor[rgb]{ .925,  .925,  .925}0.985  & \cellcolor[rgb]{ .925,  .925,  .925}\textbf{1.236 } & \cellcolor[rgb]{ .925,  .925,  .925}\textbf{0.950 } & \cellcolor[rgb]{ .925,  .925,  .925}1.456  & \cellcolor[rgb]{ .925,  .925,  .925}0.898  & \cellcolor[rgb]{ .925,  .925,  .925}\textbf{1.078 } & \cellcolor[rgb]{ .925,  .925,  .925}\textbf{0.795 } & \cellcolor[rgb]{ .925,  .925,  .925}0.527  & \cellcolor[rgb]{ .925,  .925,  .925}0.453  & \cellcolor[rgb]{ .925,  .925,  .925}\textbf{0.507 } & \cellcolor[rgb]{ .925,  .925,  .925}\textbf{0.450 } \\
    \midrule
    \multirow{5}[2]{*}{PEMS08} & 96    & 2.204  & 1.189  & \textbf{1.097 } & \textbf{0.828 } & 2.186  & 1.185  & \textbf{1.021 } & \textbf{0.789 } & 2.075  & 1.243  & \textbf{1.754 } & \textbf{1.135 } & 1.881  & 1.094  & \textbf{0.955 } & \textbf{0.789 } & 1.403  & 0.910  & \textbf{1.329 } & \textbf{0.887 } \\
          & 192   & 1.028  & 0.768  & \textbf{0.786 } & \textbf{0.686 } & 1.192  & 0.840  & \textbf{0.898 } & \textbf{0.736 } & 1.147  & 0.912  & \textbf{1.107 } & \textbf{0.898 } & \textbf{0.232 } & \textbf{0.288 } & 0.248  & 0.343  & \textbf{0.167 } & \textbf{0.255 } & 0.169  & 0.260  \\
          & 336   & 0.987  & 0.743  & \textbf{0.855 } & \textbf{0.710 } & 1.068  & 0.781  & \textbf{0.951 } & \textbf{0.752 } & 1.067  & 0.876  & \textbf{1.045 } & \textbf{0.864 } & 1.719  & 1.023  & \textbf{1.431 } & \textbf{0.956 } & \textbf{0.183 } & \textbf{0.271 } & 0.184  & 0.275  \\
          & 720   & 0.620  & 0.562  & \textbf{0.569 } & \textbf{0.537 } & 0.684  & \textbf{0.589 } & \textbf{0.676 } & 0.593  & 1.088  & 0.888  & \textbf{1.079 } & \textbf{0.884 } & 1.712  & 1.030  & \textbf{1.585 } & \textbf{1.008 } & \textbf{0.191 } & \textbf{0.282 } & 0.192  & 0.285  \\
          & \cellcolor[rgb]{ .925,  .925,  .925}Avg & \cellcolor[rgb]{ .925,  .925,  .925}1.210  & \cellcolor[rgb]{ .925,  .925,  .925}0.815  & \cellcolor[rgb]{ .925,  .925,  .925}\textbf{0.827 } & \cellcolor[rgb]{ .925,  .925,  .925}\textbf{0.690 } & \cellcolor[rgb]{ .925,  .925,  .925}1.283  & \cellcolor[rgb]{ .925,  .925,  .925}0.849  & \cellcolor[rgb]{ .925,  .925,  .925}\textbf{0.887 } & \cellcolor[rgb]{ .925,  .925,  .925}\textbf{0.717 } & \cellcolor[rgb]{ .925,  .925,  .925}1.344  & \cellcolor[rgb]{ .925,  .925,  .925}0.980  & \cellcolor[rgb]{ .925,  .925,  .925}\textbf{1.246 } & \cellcolor[rgb]{ .925,  .925,  .925}\textbf{0.945 } & \cellcolor[rgb]{ .925,  .925,  .925}1.386  & \cellcolor[rgb]{ .925,  .925,  .925}0.859  & \cellcolor[rgb]{ .925,  .925,  .925}\textbf{1.055 } & \cellcolor[rgb]{ .925,  .925,  .925}\textbf{0.774 } & \cellcolor[rgb]{ .925,  .925,  .925}0.486  & \cellcolor[rgb]{ .925,  .925,  .925}0.430  & \cellcolor[rgb]{ .925,  .925,  .925}\textbf{0.468 } & \cellcolor[rgb]{ .925,  .925,  .925}\textbf{0.427 } \\
    \midrule
    \multirow{5}[2]{*}{Solar} & 96    & 1.257  & 0.699  & \textbf{0.786 } & \textbf{0.583 } & 1.277  & 0.685  & \textbf{0.697 } & \textbf{0.530 } & 1.071  & 0.782  & \textbf{0.932 } & \textbf{0.740 } & 1.440  & 0.929  & \textbf{0.962 } & \textbf{0.743 } & 0.907  & 0.594  & \textbf{0.854 } & \textbf{0.581 } \\
          & 192   & 0.931  & \textbf{0.529 } & \textbf{0.733 } & 0.588  & 0.893  & \textbf{0.497 } & \textbf{0.716 } & 0.577  & 0.848  & 0.728  & \textbf{0.812 } & \textbf{0.715 } & 0.915  & \textbf{0.522 } & \textbf{0.752 } & 0.596  & 0.372  & \textbf{0.314 } & \textbf{0.370 } & 0.323  \\
          & 336   & 0.980  & \textbf{0.566 } & \textbf{0.872 } & 0.635  & 0.948  & \textbf{0.534 } & \textbf{0.834 } & 0.619  & 0.850  & 0.732  & \textbf{0.830 } & \textbf{0.730 } & 1.183  & 0.907  & \textbf{1.104 } & \textbf{0.856 } & 0.390  & \textbf{0.330 } & \textbf{0.388 } & 0.337  \\
          & 720   & 0.812  & \textbf{0.487 } & \textbf{0.732 } & 0.539  & 0.847  & \textbf{0.493 } & \textbf{0.769 } & 0.583  & 0.824  & \textbf{0.729 } & \textbf{0.824 } & 0.737  & 1.211  & 0.928  & \textbf{1.093 } & \textbf{0.859 } & \textbf{0.339 } & \textbf{0.318 } & 0.341  & 0.326  \\
          & \cellcolor[rgb]{ .925,  .925,  .925}Avg & \cellcolor[rgb]{ .925,  .925,  .925}0.995  & \cellcolor[rgb]{ .925,  .925,  .925}\textbf{0.570 } & \cellcolor[rgb]{ .925,  .925,  .925}\textbf{0.781 } & \cellcolor[rgb]{ .925,  .925,  .925}0.586  & \cellcolor[rgb]{ .925,  .925,  .925}0.991  & \cellcolor[rgb]{ .925,  .925,  .925}\textbf{0.552 } & \cellcolor[rgb]{ .925,  .925,  .925}\textbf{0.754 } & \cellcolor[rgb]{ .925,  .925,  .925}0.577  & \cellcolor[rgb]{ .925,  .925,  .925}0.898  & \cellcolor[rgb]{ .925,  .925,  .925}0.743  & \cellcolor[rgb]{ .925,  .925,  .925}\textbf{0.850 } & \cellcolor[rgb]{ .925,  .925,  .925}\textbf{0.731 } & \cellcolor[rgb]{ .925,  .925,  .925}1.187  & \cellcolor[rgb]{ .925,  .925,  .925}0.822  & \cellcolor[rgb]{ .925,  .925,  .925}\textbf{0.978 } & \cellcolor[rgb]{ .925,  .925,  .925}\textbf{0.763 } & \cellcolor[rgb]{ .925,  .925,  .925}0.502  & \cellcolor[rgb]{ .925,  .925,  .925}\textbf{0.389 } & \cellcolor[rgb]{ .925,  .925,  .925}\textbf{0.488 } & \cellcolor[rgb]{ .925,  .925,  .925}0.392  \\
    \bottomrule
    \end{tabular}%
  }
  \label{tab:fine_tuned_memts}
\end{table*}

\begin{figure}[htbp]
    \centering
    % --- 第一张图：ECL (Energy) ---
    \includegraphics[width=\linewidth]{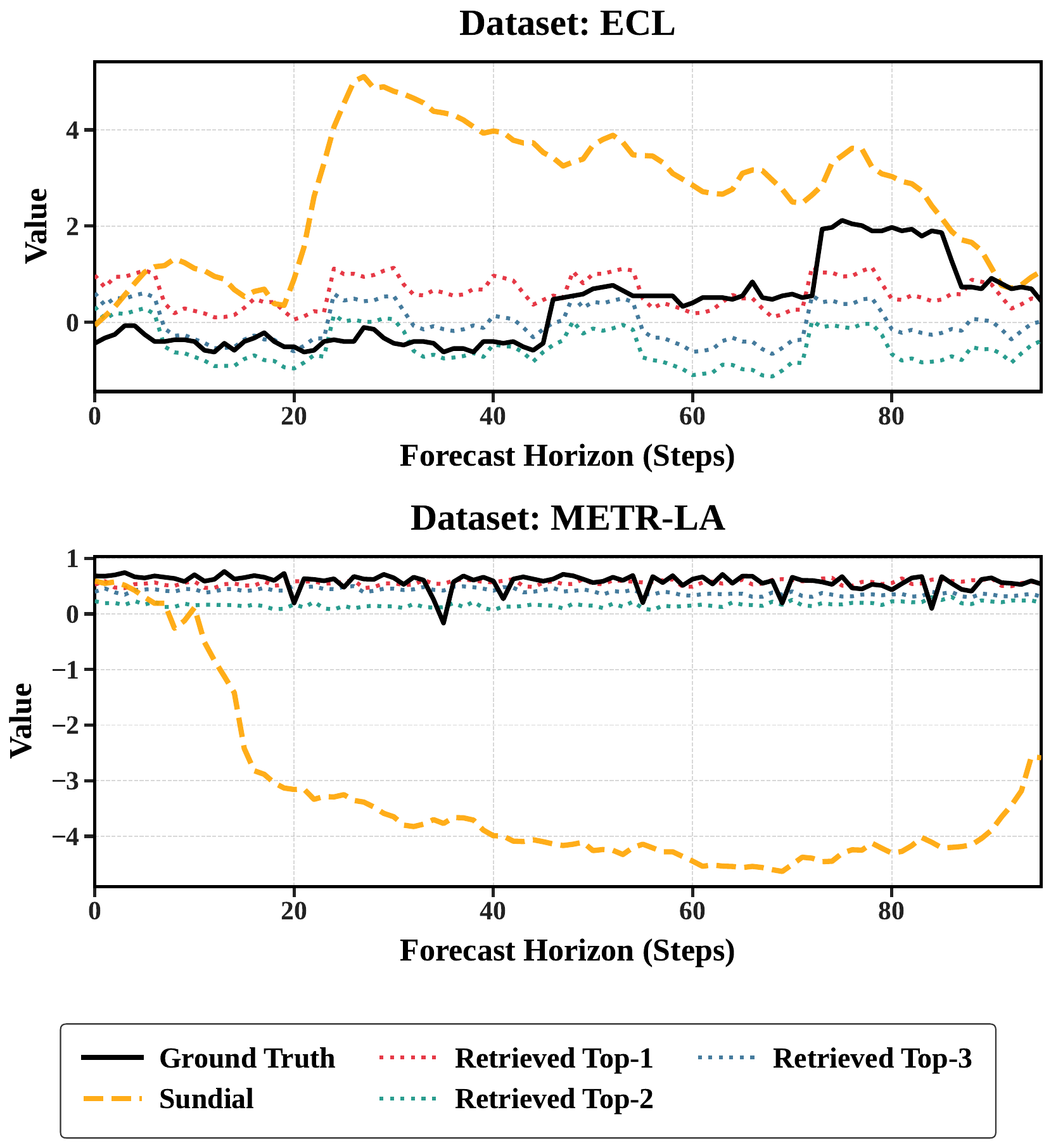}
    
    \caption{\textbf{Qualitative Analysis.} 
    (a) On ECL, MEMTS (dotted) accurately captures the peak amplitude that the Base model (dashed) underestimates. 
    (b) On METR-LA, MEMTS successfully retrieves the sharp traffic fluctuation trend, correcting the lag observed in the Base model.}
    \label{fig:vis_representative}
\end{figure}

\section{Detailed Fine-tuning Augmentation Results}
\label{app:ft_detailed}

In Section \ref{sec:ft_results} of the main text, we summarized the performance of MEMTS when integrated with fine-tuned TSFMs. Here, we provide the complete breakdown of these results. Table \ref{tab:fine_tuned_memts}  illustrates that MEMTS provides consistent gains not only on average but also across specific prediction horizons ($H \in \{96, 192, 336, 720\}$). Notably, the performance gap between ``Fine-Tuning Only'' and ``Fine-Tuning + MEMTS'' tends to widen as the prediction horizon increases, suggesting that the retrieved prototypes provide stable long-term references that prevent the fine-tuned model from overfitting to short-term noise.

\section{Visualization of Memory Retrieval Process}
\label{app:retrieve_vis}

To explicitly demonstrate the working mechanism of the Knowledge Persistence Module (KPM), we visualize the retrieval process in Figure \ref{fig:vis_representative}. The visualization highlights how the retrieved memory prototypes correct the predictions of the base foundation model.

\paragraph{Mechanism Analysis.}
As shown in the representative cases from ECL (Energy) and METR-LA (Traffic), the base foundation model (represented by the orange dashed line) often struggles with amplitude mismatch or trend lag due to distribution shifts. However, the KPM module successfully retrieves Top-$k$ historical prototypes (colored dotted lines) that exhibit high structural similarity to the Ground Truth (black solid line).
Specifically:
\begin{itemize}
    \item On the \textbf{ECL dataset} (Figure \ref{fig:vis_representative}a), the base model underestimates the peak demand. The retrieved memory sequences, however, accurately capture the high-load patterns, enabling MEMTS to adjust the forecast amplitude upwards.
    \item On the \textbf{METR-LA dataset} (Figure \ref{fig:vis_representative}b), the base model fails to predict the sharp drop in traffic speed. In contrast, the retrieved prototypes clearly indicate a congestion pattern, allowing MEMTS to correct the trend and align closely with the ground truth.
\end{itemize}
These examples confirm that the performance gain of MEMTS stems from its ability to recall and utilize domain-specific "future candidates" that are missing from the generalist backbone's context.

\section{Visual Benchmarking across Diverse Domains}
\label{app:zero_shot_vis}

\begin{figure*}[t]
    \centering
    \begin{minipage}{0.85\textwidth}
        \centering
        \begin{subfigure}[b]{0.32\textwidth}
            \centering
            \includegraphics[width=\textwidth]{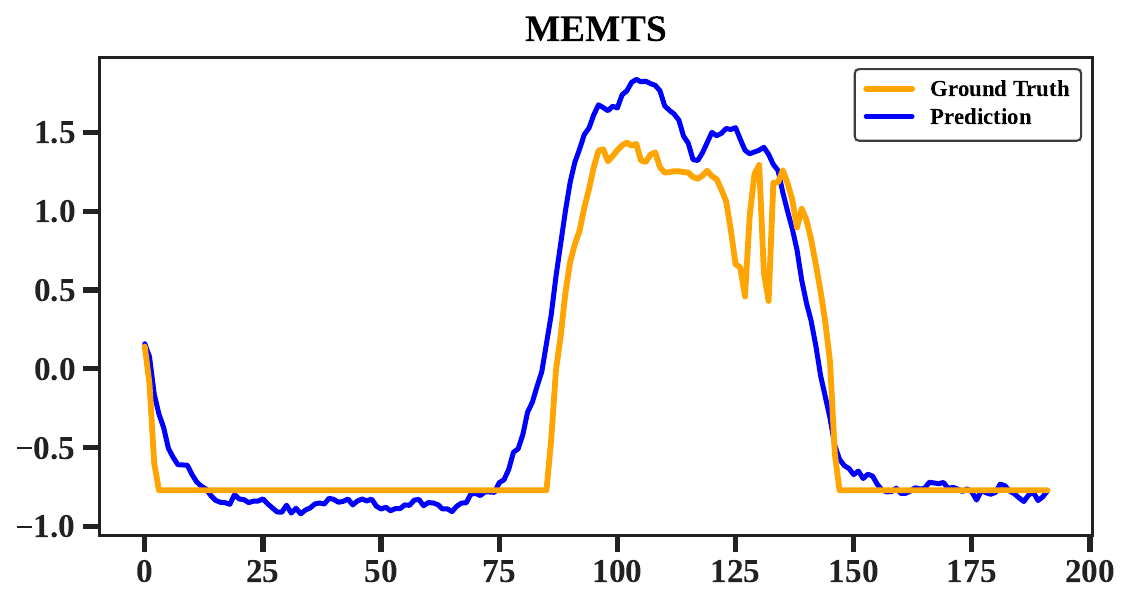}
            \label{fig:Solar-memts}
        \end{subfigure}
        \hspace{-0.02\textwidth}
        \begin{subfigure}[b]{0.32\textwidth}
            \centering
            \includegraphics[width=\textwidth]{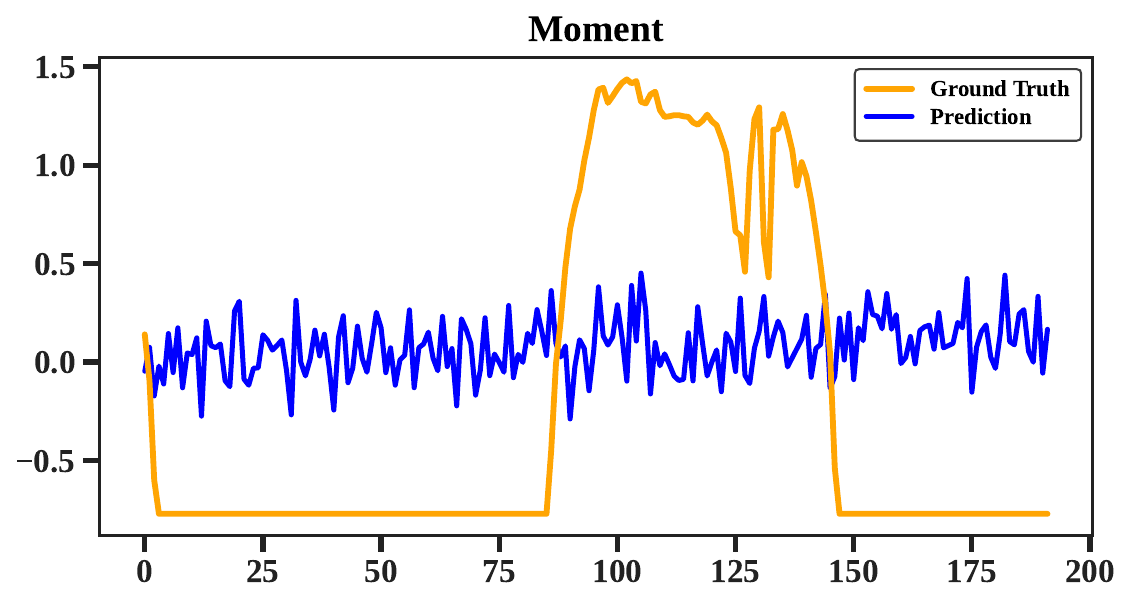}
            \label{fig:Solar-moment}
        \end{subfigure}
        \hspace{-0.02\textwidth}
        \begin{subfigure}[b]{0.32\textwidth}
            \centering
            \includegraphics[width=\textwidth]{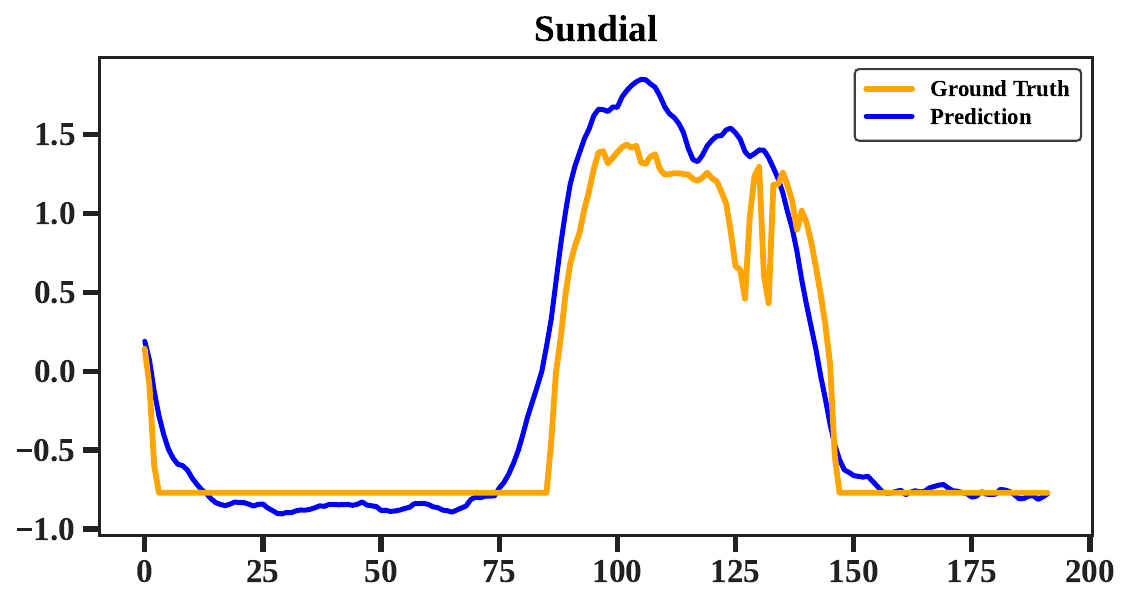}
            \label{fig:Solar-sundial}
        \end{subfigure}
        
        \vspace{0.1cm}
        
        % 第二行
        \begin{subfigure}[b]{0.32\textwidth}
            \centering
            \includegraphics[width=\textwidth]{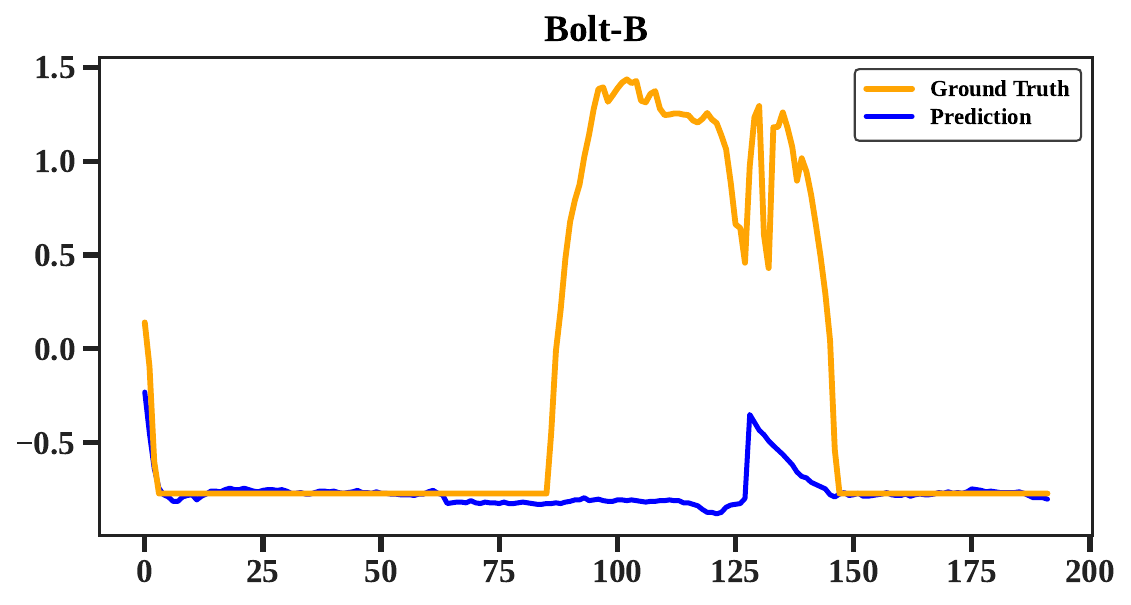}
            \label{fig:Solar-chronos_base}
        \end{subfigure}
        \hspace{-0.02\textwidth}
        \begin{subfigure}[b]{0.32\textwidth}
            \centering
            \includegraphics[width=\textwidth]{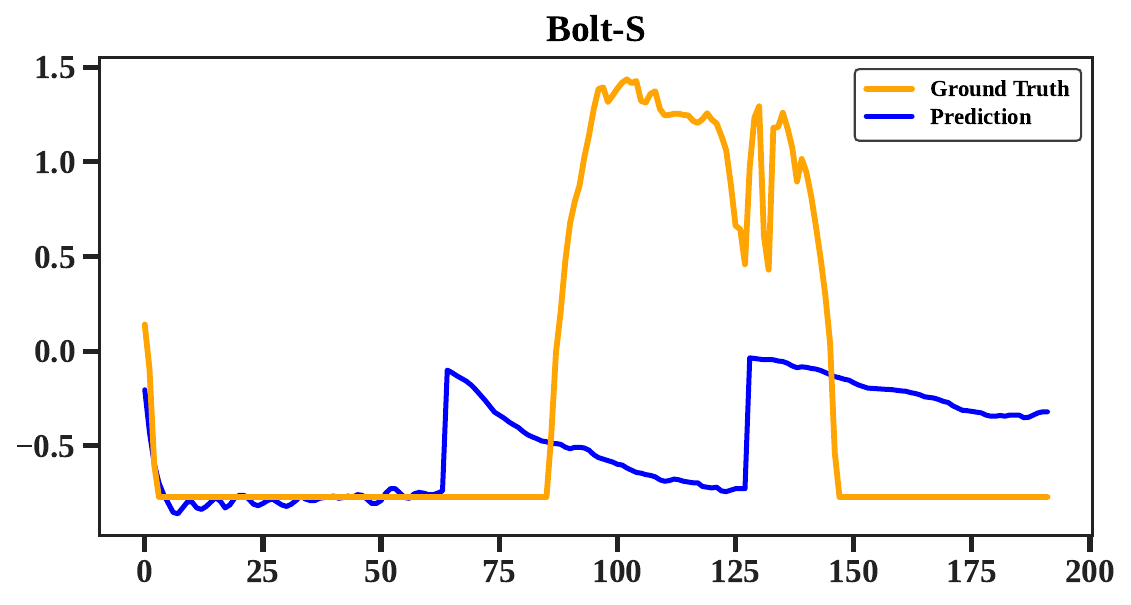}
            \label{fig:Solar-chronos_small}
        \end{subfigure}
        \hspace{-0.02\textwidth}
        \begin{subfigure}[b]{0.32\textwidth}
            \centering
            \includegraphics[width=\textwidth]{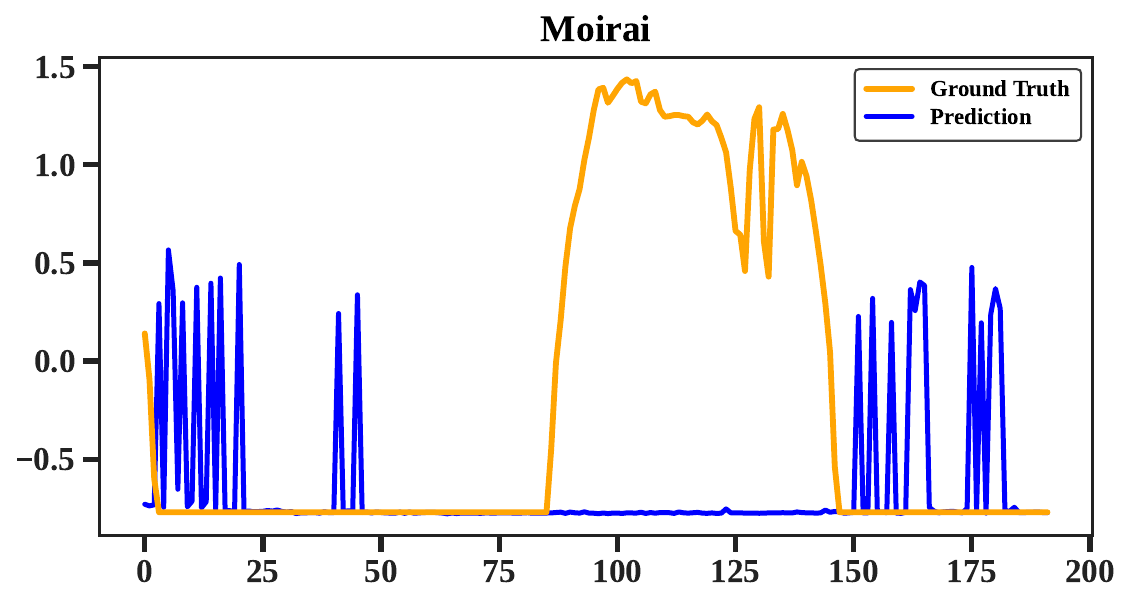}
            \label{fig:Solar-moirai}
        \end{subfigure}
        \caption{Visualization of input-336-predict-192 results on the Solar dataset.}
        \label{fig:Solar}
    \end{minipage} % 结束minipage
\end{figure*}

\begin{figure*}[t]
    \centering
    \begin{minipage}{0.85\textwidth}
        \centering
        \begin{subfigure}[b]{0.32\textwidth}
            \centering
            \includegraphics[width=\textwidth]{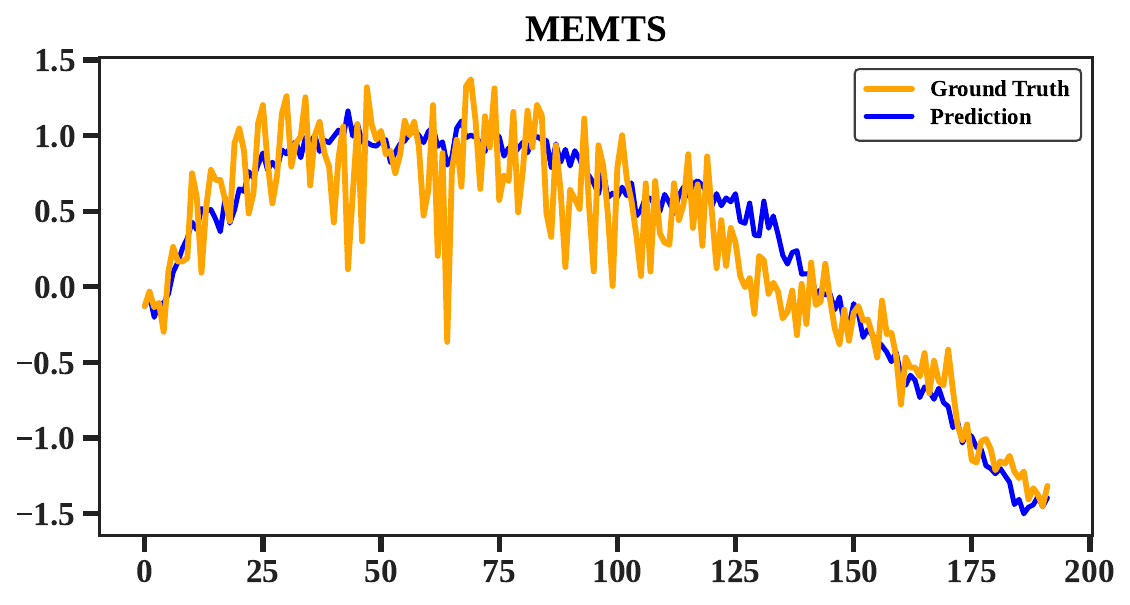}
            \label{fig:PEMS04-memts}
        \end{subfigure}
        \hspace{-0.02\textwidth}
        \begin{subfigure}[b]{0.32\textwidth}
            \centering
            \includegraphics[width=\textwidth]{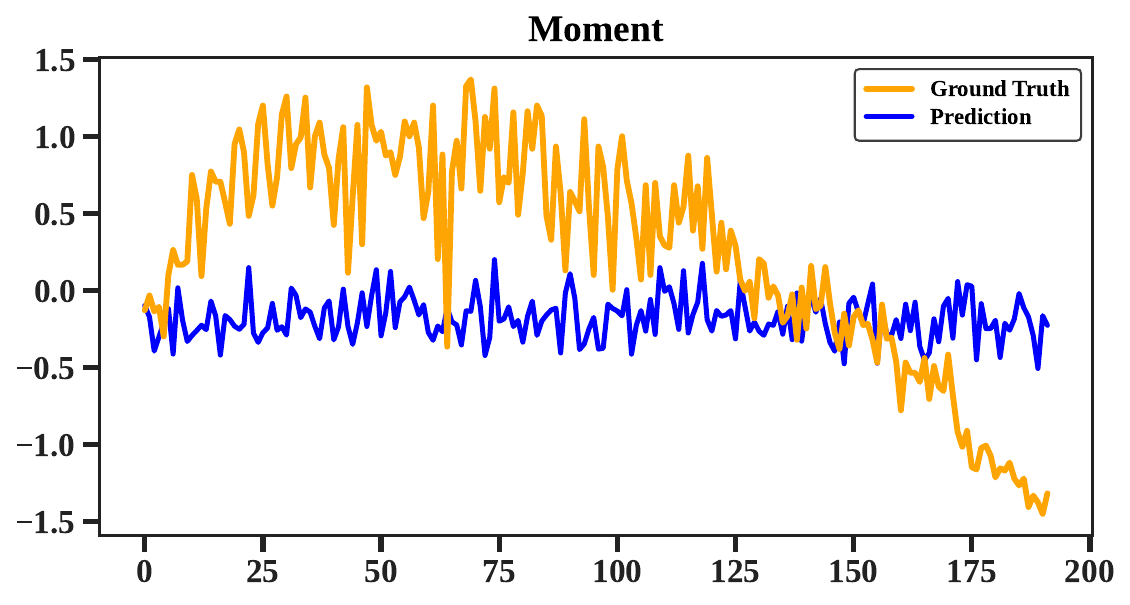}
            \label{fig:PEMS04-moment}
        \end{subfigure}
        \hspace{-0.02\textwidth}
        \begin{subfigure}[b]{0.32\textwidth}
            \centering
            \includegraphics[width=\textwidth]{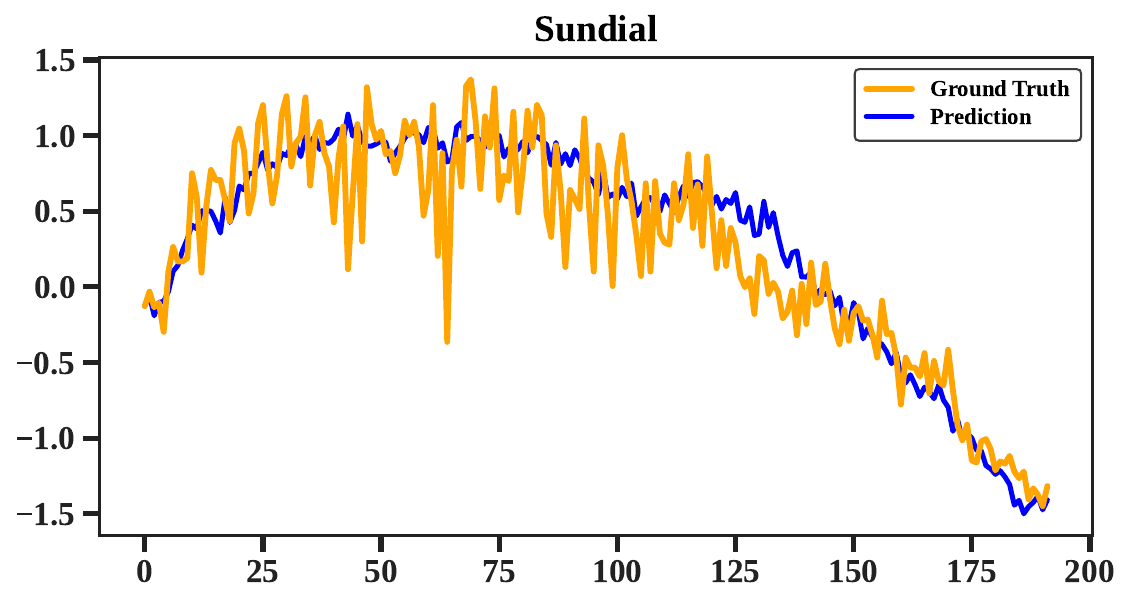}
            \label{fig:PEMS04-sundial}
        \end{subfigure}
        
        \vspace{0.1cm}
        
        % 第二行
        \begin{subfigure}[b]{0.32\textwidth}
            \centering
            \includegraphics[width=\textwidth]{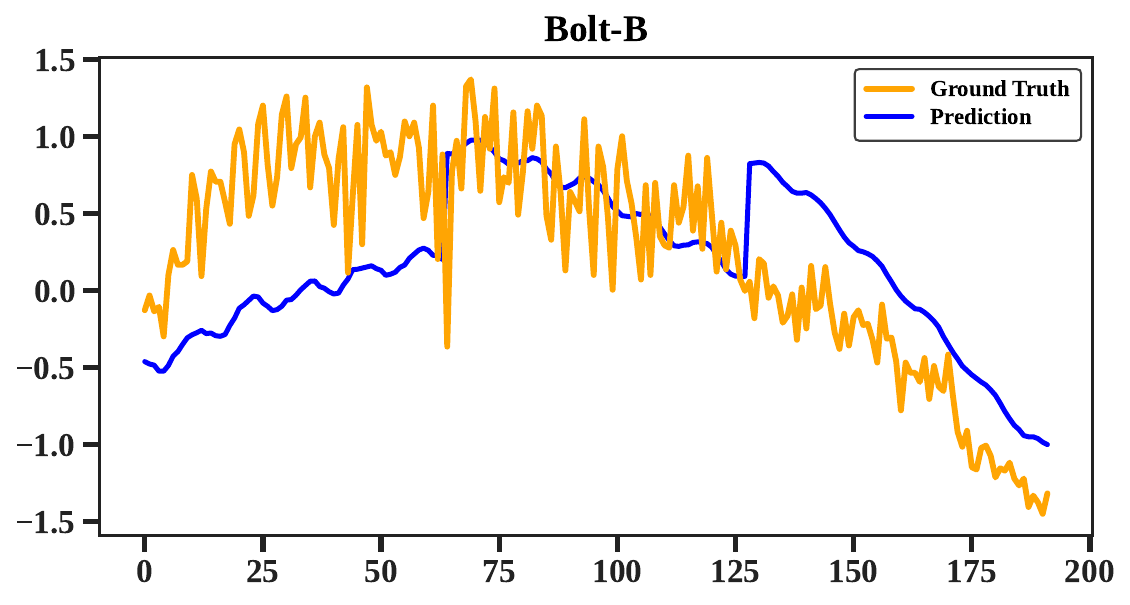}
            \label{fig:PEMS04-chronos_base}
        \end{subfigure}
        \hspace{-0.02\textwidth}
        \begin{subfigure}[b]{0.32\textwidth}
            \centering
            \includegraphics[width=\textwidth]{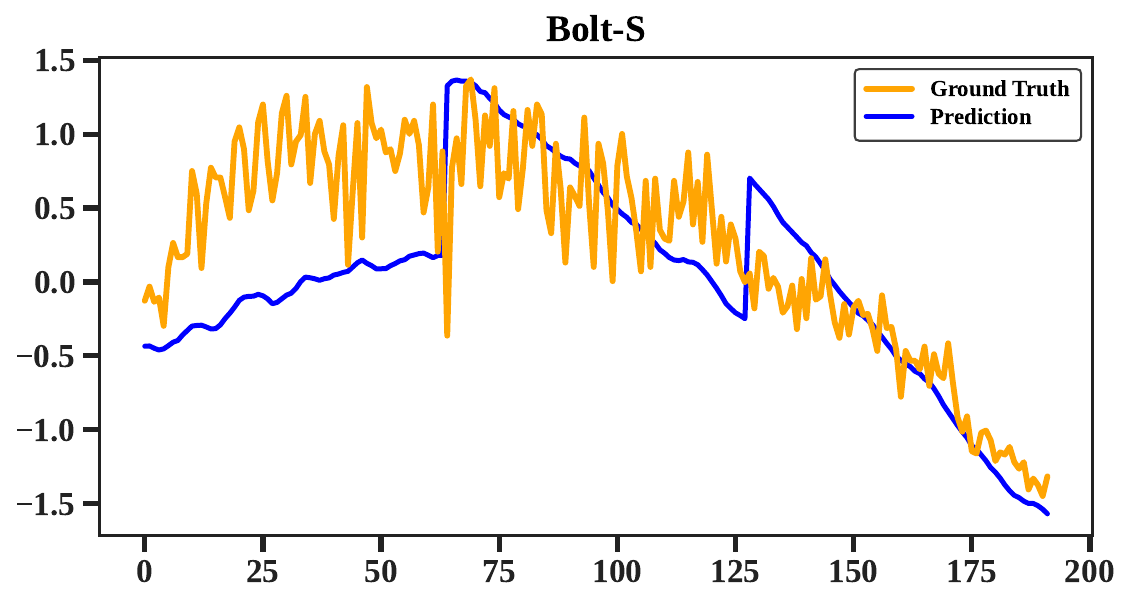}
            \label{fig:PEMS04-chronos_small}
        \end{subfigure}
        \hspace{-0.02\textwidth}
        \begin{subfigure}[b]{0.32\textwidth}
            \centering
            \includegraphics[width=\textwidth]{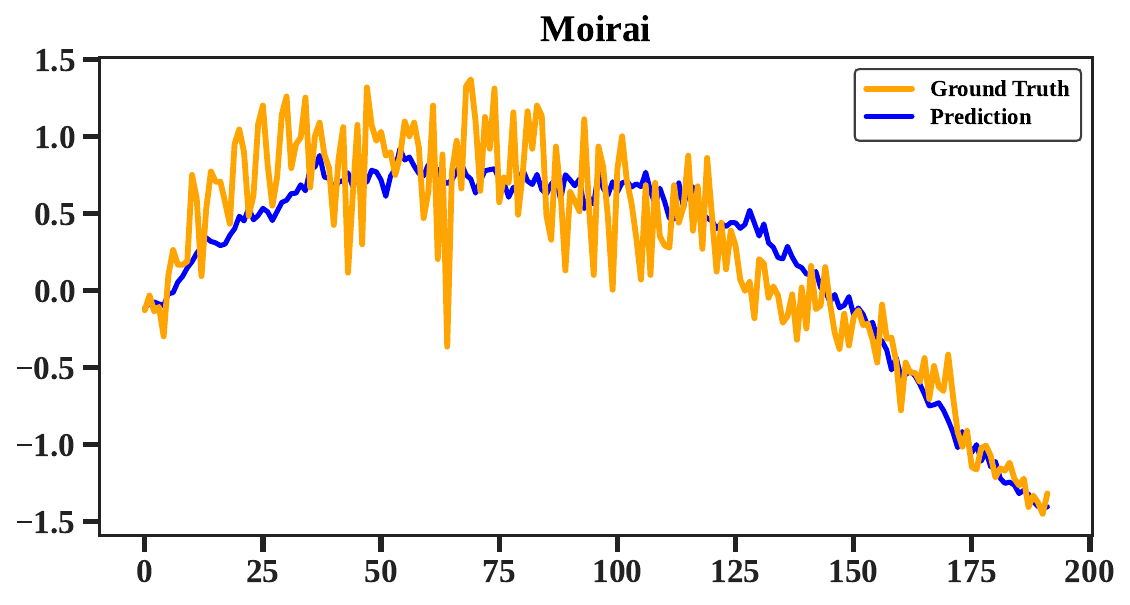}
            \label{fig:PEMS04-moirai}
        \end{subfigure}
        \caption{Visualization of input-336-predict-192 results on the PEMS04 dataset.}
        \label{fig:PEMS04}
    \end{minipage} % 结束minipage
\end{figure*}

\begin{figure*}[t]
    \centering
    \begin{minipage}{0.85\textwidth}
        \centering
        \begin{subfigure}[b]{0.32\textwidth}
            \centering
            \includegraphics[width=\textwidth]{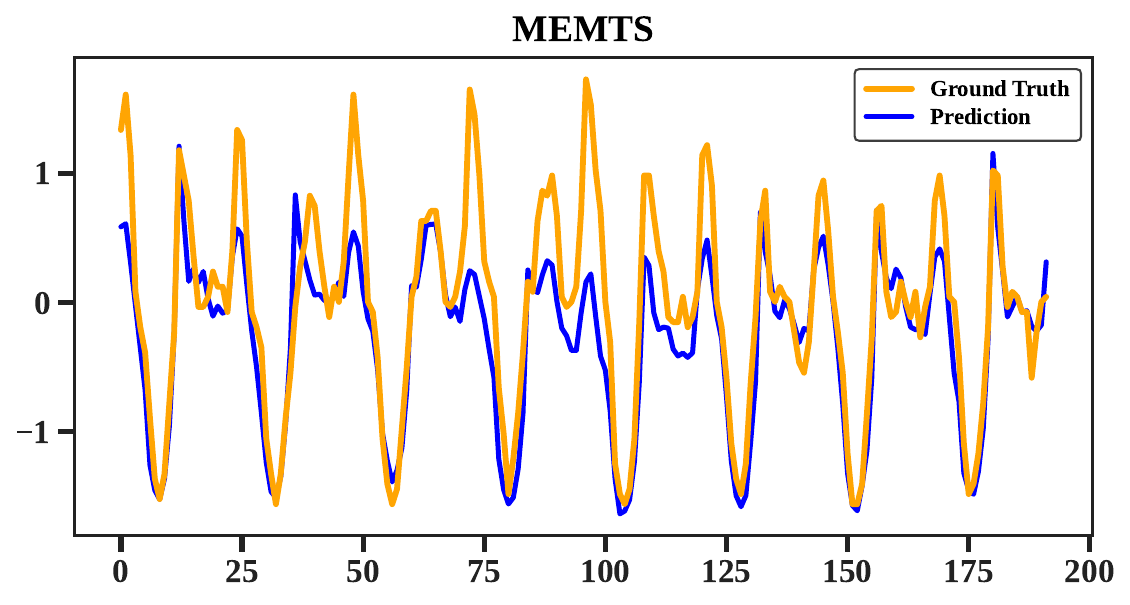}
            \label{fig:ecl-memts}
        \end{subfigure}
        \hspace{-0.02\textwidth}
        \begin{subfigure}[b]{0.32\textwidth}
            \centering
            \includegraphics[width=\textwidth]{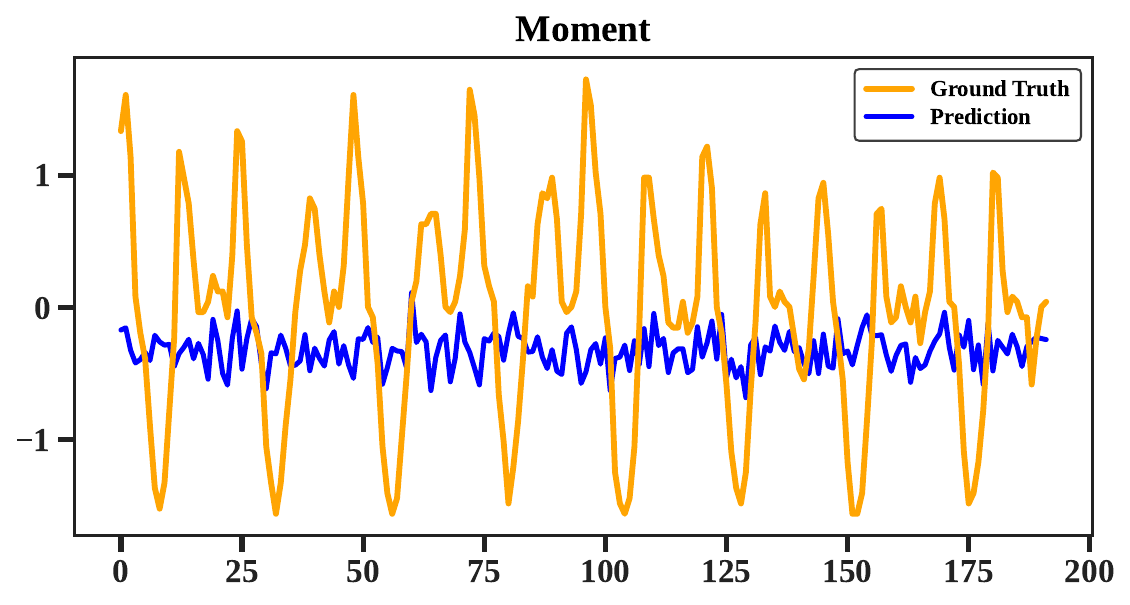}
            \label{fig:ecl-moment}
        \end{subfigure}
        \hspace{-0.02\textwidth}
        \begin{subfigure}[b]{0.32\textwidth}
            \centering
            \includegraphics[width=\textwidth]{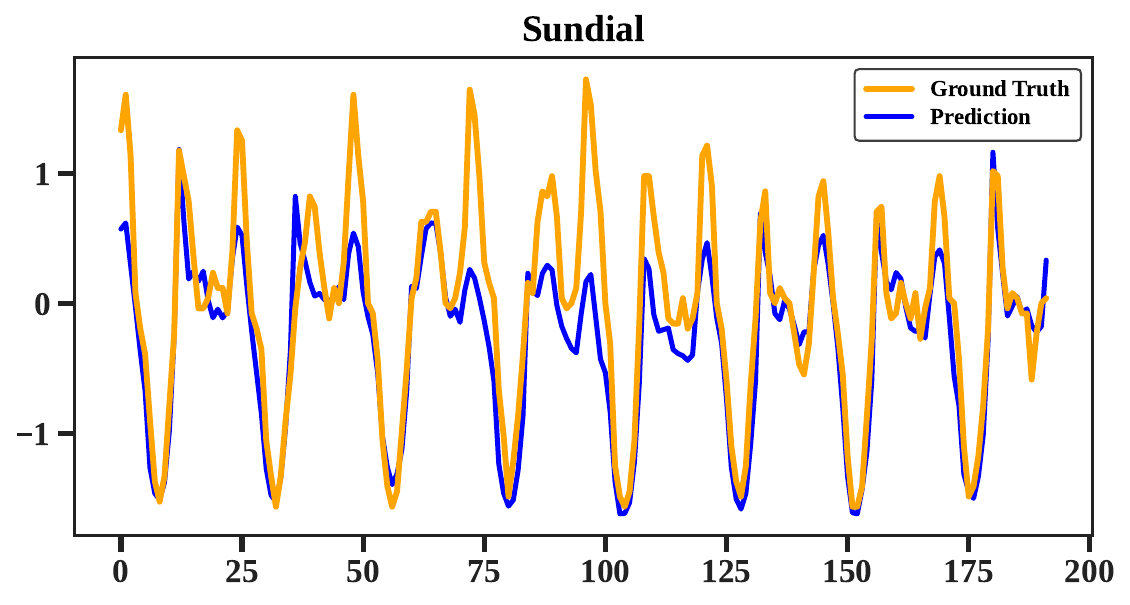}
            \label{fig:ecl-sundial}
        \end{subfigure}
        
        \vspace{0.1cm}
        
        % 第二行
        \begin{subfigure}[b]{0.32\textwidth}
            \centering
            \includegraphics[width=\textwidth]{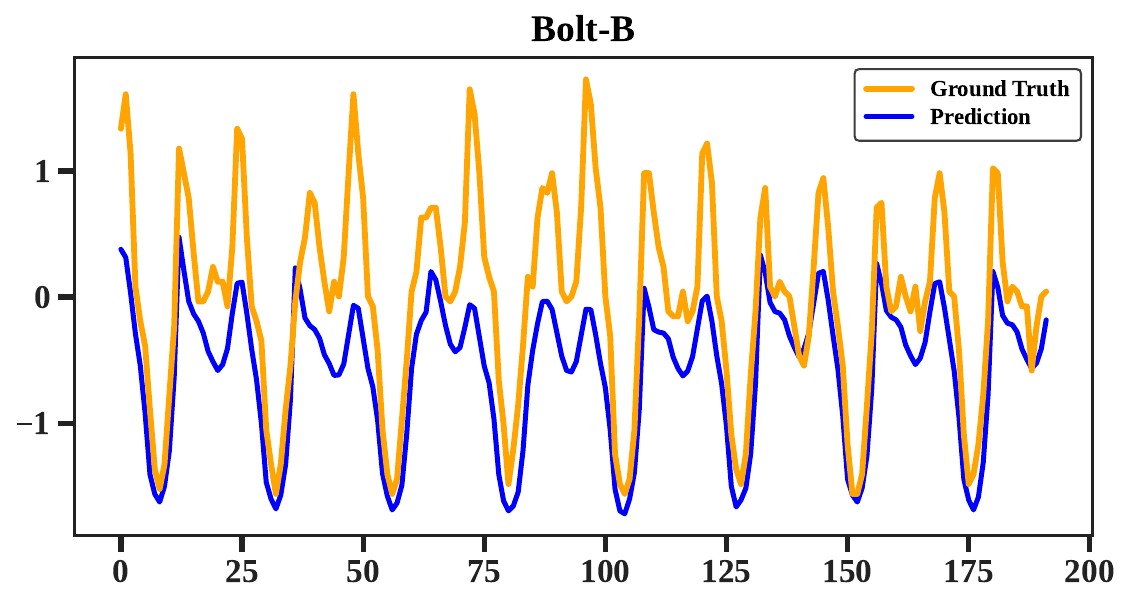}
            \label{fig:ecl-chronos_base}
        \end{subfigure}
        \hspace{-0.02\textwidth}
        \begin{subfigure}[b]{0.32\textwidth}
            \centering
            \includegraphics[width=\textwidth]{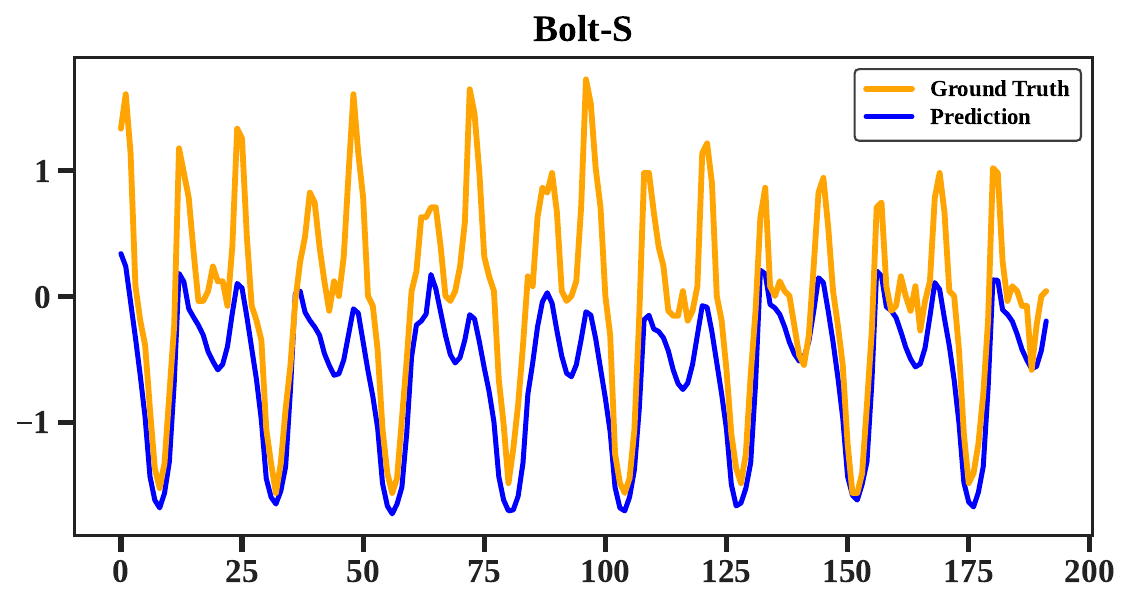}
            \label{fig:ecl-chronos_small}
        \end{subfigure}
        \hspace{-0.02\textwidth}
        \begin{subfigure}[b]{0.32\textwidth}
            \centering
            \includegraphics[width=\textwidth]{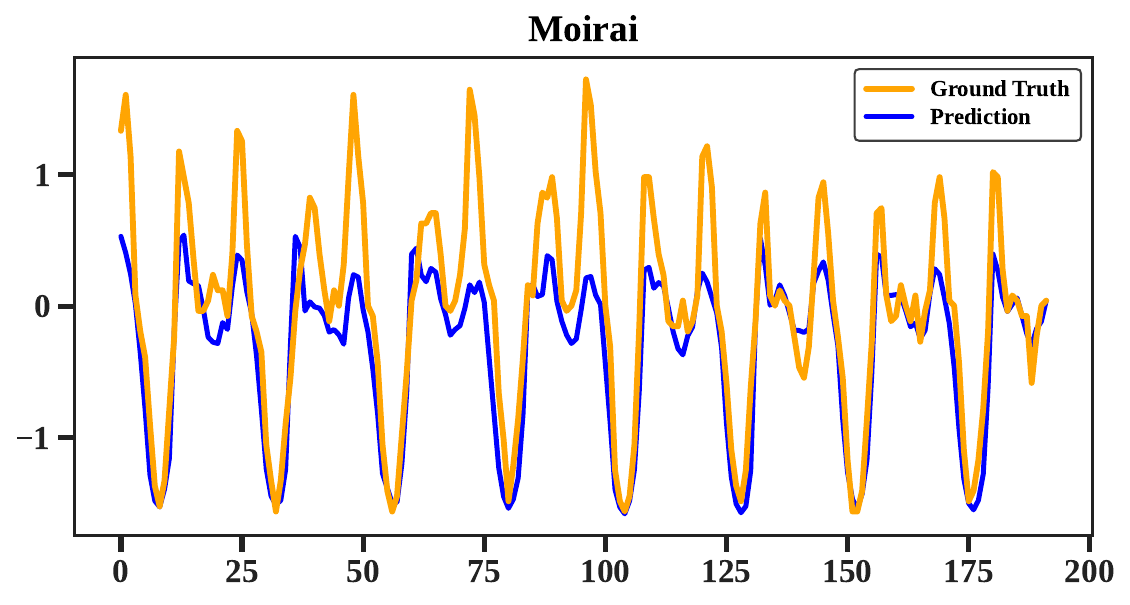}
            \label{fig:ecl-moirai}
        \end{subfigure}
        \caption{Visualization of input-336-predict-192 results on the ECL dataset.}
        \label{fig:ecl}
    \end{minipage} % 结束minipage
\end{figure*}

As illustrated in Figures \ref{fig:Solar}, \ref{fig:PEMS04}, and \ref{fig:ecl}, we present a comprehensive visual comparison between MEMTS and five state-of-the-art Time Series Foundation Models (Moment, Sundial, Bolt-B, Bolt-S, and Moirai). The evaluation is conducted under a zero-shot setting with a historical context length of $L=336$ and a prediction horizon of $H=192$. Covering representative datasets from the energy (Solar, ECL) and traffic (PEMS04) domains, the visualizations clearly demonstrate that MEMTS consistently yields superior forecasting fidelity. While baseline models often exhibit noise or trend misalignment, MEMTS generates smooth and accurate predictions that closely align with the ground truth, validating its robustness across different data distributions.

\end{document}